\documentclass[10pt]{article} % For LaTeX2e
\usepackage[T1]{fontenc}
\usepackage{mathptmx}
\usepackage[preprint]{tmlr}

\usepackage{graphicx}
\usepackage{subcaption} % For subfigures/subtables
\usepackage{amsmath}
\usepackage{amssymb}
\usepackage{booktabs}
\usepackage{url}
\usepackage{xcolor}         % colors
\usepackage{xspace}
\graphicspath{ {figures/} }
\usepackage{booktabs} % for professional tables
\usepackage{enumitem}
\usepackage{multirow}
\usepackage[normalem]{ulem}

\usepackage[pagebackref,breaklinks,colorlinks]{hyperref}
\usepackage{url}

\hypersetup{
  colorlinks,
  breaklinks=true,
  anchorcolor={blue!50!black},
  linkcolor={blue!50!black},
  citecolor={blue!50!black},
  urlcolor={blue!50!black}
}

\usepackage[capitalize]{cleveref}
\crefname{section}{Appendix}{Secs.}
\Crefname{section}{Section}{Sections}
\Crefname{table}{Table}{Tables}
\crefname{table}{Tab.}{Tabs.}

\newcommand{\vx}{\mathbf{x}}
\newcommand{\vy}{\mathbf{y}}
\newcommand{\vm}{\mathbf{m}}
\newcommand{\thetamin}{\theta_{\min}}

\usepackage{amsmath,amsfonts,bm}

\def\eqref#1{equation~\ref{#1}}
\def\1{\bm{1}}
\def\vtheta{{\bm{\theta}}}

\def\vm{{\bm{m}}}

\def\vx{{\bm{x}}}
\def\vy{{\bm{y}}}

\DeclareMathAlphabet{\mathsfit}{\encodingdefault}{\sfdefault}{m}{sl}
\SetMathAlphabet{\mathsfit}{bold}{\encodingdefault}{\sfdefault}{bx}{n}

\usepackage{amsmath,amsfonts,bm}

\def\eqref#1{equation~\ref{#1}}
\def\1{\bm{1}}
\def\vtheta{{\bm{\theta}}}

\def\vm{{\bm{m}}}

\def\vx{{\bm{x}}}
\def\vy{{\bm{y}}}

\DeclareMathAlphabet{\mathsfit}{\encodingdefault}{\sfdefault}{m}{sl}
\SetMathAlphabet{\mathsfit}{bold}{\encodingdefault}{\sfdefault}{bx}{n}

\renewcommand{\eqref}[1]{Eq.~(\ref{#1})}

\def\vy{{\bm{y}}}

\renewcommand{\paragraph}[1]{\noindent\textbf{#1}}

\usepackage{authblk}
\title{Oya: Deep Learning for Accurate Global Precipitation \\ Estimation}
\author[1]{Emmanuel Asiedu Brempong}
\author[1]{Mohammed Alewi Hassen}
\author[1]{MohamedElfatih MohamedKhair}
\author[1]{Vusumuzi Dube}
\author[1]{Santiago Hincapie Potes}
\author[1]{Olivia Graham}
\author[1]{Amanie Brik}
\author[2]{Amy McGovern}
\author[3]{George J. Huffman}
\author[1]{Jason Hickey}
\affil[1]{Google Research Africa}
\affil[2]{University of Oklahoma}
\affil[3]{NASA Goddard Space Flight Center}

\begin{document}

\maketitle

\footnotetext{Corresponding author(s): brempong@google.com}

\vspace{-.2cm}
\begin{abstract}

Accurate precipitation estimation is critical for hydrological applications, especially in the Global South where ground-based observation networks are sparse and forecasting skill is limited. Existing satellite-based precipitation products often rely on the longwave infrared 
% ($10.7 \mu m$) 
channel alone or are calibrated with data that can introduce significant errors, particularly at sub-daily timescales. This study introduces Oya, a novel real-time precipitation retrieval algorithm utilizing the full spectrum of visible and infrared (VIS-IR) observations from geostationary (GEO) satellites. Oya employs a two-stage deep learning approach, combining two U-Net models: one for precipitation detection and another for quantitative precipitation estimation (QPE), to address the inherent data imbalance between rain and no-rain events. The models are trained using high-resolution GPM Combined Radar-Radiometer Algorithm (CORRA) v07 data as ground truth and pre-trained on IMERG-Final retrievals to enhance robustness and mitigate overfitting due to the limited temporal sampling of CORRA. By leveraging multiple GEO satellites, Oya achieves quasi-global coverage and demonstrates superior performance compared to existing competitive regional and global precipitation baselines, offering a promising pathway to improved precipitation monitoring and forecasting.
 
\end{abstract}

\section{Introduction}

Precipitation is a fundamental driver of hydrological processes, impacting agriculture, water resource management, and disaster preparedness. Accurate and reliable precipitation estimates are crucial, particularly in the Global South, where rain-fed agriculture is prevalent and vulnerability to climate extremes is high~\citep{fao2017state}. Errors in precipitation estimates propagate through forecasting systems, compromising the reliability of subsequent predictions. This cascading effect underscores the critical need for accurate precipitation observations as a foundation for effective forecasting.

Traditional methods of precipitation measurement, primarily relying on ground-based radar and rain gauges, face significant limitations in the Global South.
For example in Africa, most countries, with the notable exception of South Africa, do not have a single working radar station.
Rain gauges can offer precise measurements but their scarcity across the Global South poses a significant challenge. The density of reporting weather stations in these regions falls drastically short of the World Meteorological Organization's (WMO) recommendations and is, alarmingly, on the decline~\citep{ComparisonofRainfallProductsoverSubSaharanAfrica, MEKONNEN2023101514}.
As such, the quality of both precipitation forecasts and analyses are affected. 
% For example, single-day tropical precipitation forecasts had similar skill to a six day forecast for the extratropics~\citep{IntercomparisonofGlobalModelPrecipitationForecastSkillin201011UsingtheSEEPSScore}.
In addition, tropical precipitation is often driven by mesoscale effects like deep convection, making precipitation forecasts and analyses intrisically more difficult than in the extratopics.
% As a result, single-day tropical precipitation forecasts, for instance, had similar skill to a six day forecast for the extratropics~\citep{IntercomparisonofGlobalModelPrecipitationForecastSkillin201011UsingtheSEEPSScore}.

Satellite-based precipitation estimates provide a valuable and increasingly skillful alternative to ground-based observation networks. These retrievals rely mostly on input from geostationary (GEO) satellites and/or low earth orbiting (LEO) satellites.
LEO satellites often use passive microwave sensors (PMW) to observe precipitation. PMW sensors provide reasonably direct estimates of precipitation and thus are recognized as a reliable source for instantaneous precipitation estimates~\citep{ebert1996results}.
However, owing to their low spatial and temporal density, PMW observations from individual satellites cannot easily be used to provide precipitation estimates with complete spatial coverage.
To achieve global coverage, the Global Precipitation Measurement (GPM) mission’s Integrated Multisatellite Retrievals for GPM (IMERG)~\citep{Huffman2020} merges observations from a constellation of PMW satellites.
Regions without valid PMW observations are filled in by morphing the observations from the closest satellite observations, which can significantly reduce performance~\citep{AProcessBasedValidationofGPMIMERGandItsSourcesUsingaMesoscaleRainGaugeNetworkintheWestAfricanForestZone}.
The Climate Prediction Center (CPC) Morphing algorithm (CMORPH) ~\citep{CMORPHAMethodthatProducesGlobalPrecipitationEstimatesfromPassiveMicrowaveandInfraredDataatHighSpatialandTemporalResolution} and the Global Satellite Map of Precipitation (GSMaP) ~\citep{GSMaP} datasets also employ the PMW constellation to produce precipitation estimates.

GEO satellites, in contrast, offer observations at high spatial and temporal density across multiple spectral channels. For example, the  Geostationary Operational Environmental Satellite (GOES) series of satellites operated by NOAA covering the Americas, currently have 16 spectral channels, spanning visible (VIS: 2 channels at $0.47 \mu m$ and $0.64 \mu m$), near infrared (4 channels from $0.86 \mu m$ to $2.2 \mu m$), and infrared (IR: 10 channels from $3.9 \mu m$ to $13.3 \mu m$).
Each band is sensitive to different cloud and atmospheric properties, but none of the bands provides a direct measurement of precipitation.
Thus, algorithms developed for precipitation retrievals from GEO observations are essentially based on deriving empirical relationships between the surface precipitation rate and the cloud properties measured by the satellites~\citep{Scofield1977}.
The Convective Rainfall Rate (CRR) algorithm developed within the European Organisation for the Exploitation of Meteorological Satellites (EUMETSAT) Nowcasting Satellite Applications Facility (NWCSAF)~\citep{marcos2016algorithm} is one such IR-based precipitation estimation algorithm.
It estimates precipitation rates using data from three channels (VIS ($0.6 \mu m$), longwave IR 
% ($10.7 \mu m$\footnote[1]{The exact wavelength differs slightly for each of the current generation of GEO satellites. $10.7 \mu m$ is the traditional wavelength for this band and we use that here for consistency. \cref{section:channels-summary} provides the detailed wavelength for each satellite.})
, and WV ($6.2 \mu m$)) from the Meteosat Second Generation (MSG) SEVIRI instrument.
The estimates are obtained from calibration analytical functions which are generated by combining SEVERI and radar data.
The Global Challenges Research Fund (GCRF) African Science for Weather Information and Forecasting Techniques (SWIFT) project has operationalized the use of CRR for nowcasting applications in the National Meteorological Services (NMS) of multiple African countries.
Though limited in retrieval  skill, the CRR has proved to be a valuable tool due to its free availability and near real-time operation, which are essential for nowcasting applications, particularly in areas with funding limitations ~\citep{TheAfricanSWIFTProjectGrowingScienceCapabilitytoBringaboutaRevolutioninWeatherPrediction, africanowcastwithnwcsaf}. 

Other precipitation estimation products based on the longwave IR 
% ($10.7 \mu m$)
channel alone include the Hydro-Estimator (HE)~\citep{StatusandOutlookofOperationalSatellitePrecipitationAlgorithmsforExtremePrecipitationEvents}, the  Precipitation Estimation from Remotely Sensed Information Using Artificial Neural Networks - Cloud Classification System (PERSIANN-CCS)~\citep{PrecipitationEstimationfromRemotelySensedImageryUsinganArtificialNeuralNetworkCloudClassificationSystem}, and the Self-Calibrating Multivariate Precipitation Retrieval (SCaMPR)~\citep{ASelfCalibratingRealTimeGOESRainfallAlgorithmforShortTermRainfallEstimates}. 
PERSIANN Dynamic Infrared Rain Rate (PDIR-Now) ~\citep{PDIR-Now} improves upon PERSIAN-CCS by dynamically shifting the longwave IR 
% ($10.7 \mu m$) 
brightness temperature - rain rate (Tb-R) curves at each pixel based on the relative wetness of the pixel’s climatology.

% Despite the availability of multiple channels, many operational precipitation retrieval algorithms rely on the longwave IR 

This limitation to a single channel and single pixel significantly reduces the computational complexity, making it feasible for operational implementation in prior generations of computer resources. 
% ($10.7 \mu m$)
% channel alone, linking colder cloud tops to higher precipitation rates.
Algorithms like the CRR, which uses other channels in addition to the longwave IR 
% ($10.7 \mu m$)
channel, demonstrate the potential benefit of using multiple channels for improved estimates. The study in this article builds on the premise that leveraging the full spectrum of available channels should provide more robust empirical relationships between satellite-observed cloud properties and surface precipitation rates, ultimately leading to more accurate estimates, an advantage explored further in ~\Cref{section:number-of-channels}.

Further, products like CRR and PERSIAN-CCS are based on Tb-R relationships derived based on ground radar observations.
These Tb-R relationships are then used not only for retrievals in the same regions where the ground truth is sourced, but are applied to  other regions where the ground observations are not available.
The obvious result of this is that the quality of the retrievals in regions without the requisite ground observations are lower than in regions that have them. Other products, such SCaMPR and PDIR, are calibrated against PMW precipitation estimates. 
An error budget analysis of these products show that a major portion of their errors are propagated from the PMW calibration data~\citep{OnthePropagationofSatellitePrecipitationEstimationErrorsFromPassiveMicrowavetoInfraredEstimates}.
On the daily and monthly scales and over large grid box sizes ($1^{\circ} \times 1^{\circ}$ to $5^{\circ} \times 5^{\circ}$), there is evidence that PERSIANN precipitation retrievals have high skill~\citep{EvaluationofPERSIANNSystemSatelliteBasedEstimatesofTropicalRainfall}.
For the sub-daily/instantaneous time scale, however, we find that the PDIR-Now retrievals have significantly lower quality. 

In terms of direct observations, the GPM Combined Radar-Radiometer Algorithm (CORRA)~\citep{gpm2b} provides the most accurate, high resolution estimates of surface precipitation rate that can be achieved from a spaceborne platform~\citep{TheGPMCombinedAlgorithm}. 
The GPM Core Observatory (CO) satellite uses a Dual-frequency Precipitation Radar (DPR), which scans cross-track in
relatively narrow swaths at Ku band (13.6 GHz) and Ka band (35.5 GHz). The dual-frequency radar reflectivity observations are nearly beam-matched over the 125km
Ka-band swath, with a horizontal resolution of approximately 5km, and a vertical
resolution of 250m in standard observing mode.
However, unlike PMW-based estimates, where a constellation of sensors on different platforms can be merged to achieve low revisit times, the GPM CO includes only a single instrument set with revisit times of 2.5 days.
Thus, these estimates have extremely low temporal sampling and have mainly been used as a reference dataset for cross-calibrating precipitation estimates from all of the PMW radiometers in the GPM constellation~\citep{Huffman2020}.
Their usage as ground truth for IR-based precipitation estimation algorithms has had little attention.
With recent advances in deep learning, there is an opportunity to utilise these observations to enhance the quality of precipitation estimation using machine learning models (in this article, we use the term ``model'' to refer to statistical machine learning models, rather than traditional numerical weather prediction models).
\cite{nrtestimationbrazil} demonstrated the viability of this approach with a CNN model trained using the GPM CORRA observations as targets over Brazil.
This model outperformed the operational HYDRO \citep{nrtestimationbrazil} retrievals in use at INPE in Brazil.
Other studies using ground observations as targets instead of satellite radar have also demonstrated CNNs to consistently outperform other simpler statistical models including,  PERSIANN-CNN~\citep{PERSIANNCNNPrecipitationEstimationfromRemotelySensedInformationUsingArtificialNeuralNetworksConvolutionalNeuralNetworks} and PEISCNN~\citep{PrecipitationEstimationBasedonInfraredDatawithaSphericalConvolutionalNeuralNetwork}.

In this study, we present Oya, a real-time precipitation retrieval algorithm that relies on the full spectrum of observations from GEO satellites as inputs.
The seventh version (v07) of the GPM CORRA precipitation observations are used as ground truth for training the models.
To deal with the huge imbalance in favour of no-rain observations, Oya combines the output of two UNet~\citep{unet} models, one trained only to identify precipitation events and the other a Quantitative Precipitation Estimation (QPE) model trained to estimate the actual amount of precipitation.
Owing to the aforementioned limitations of the LEO satellites, the GPM CORRA v07 observations are barely enough to train a model without overfitting.
To increase the amount of data to which the models are exposed and thus avoid overfitting, the models are pretrained on retrievals from the IMERG Final run which has complete spatial coverage around the globe.
We apply this approach across multiple GEO satellites to produce a quasi-global precipitation estimate that outperforms similar competitive regional and global baselines.

\section{Data} \label{data}

\subsection{Geostationary Satellites}

The Oya estimation model utilizes the full spectrum of observations from  GEO satellites as input data for precipitation retrieval. To produce global ($60 ^\circ$N$-60 ^\circ$S) precipitation estimates, we rely on data from a constellation of satellites.
With data from the GOES-16 and GOES-18 Advanced Baseline Imagers (ABI)~\citep{INTRODUCINGTHENEXTGENERATIONADVANCEDBASELINEIMAGERONGOESR}, we produce models to cover the Americas and with data from MSG SEVIRI instrument aboard the Meteosat-9 Indian Ocean Data Coverage (IODC service) and Meteosat-10 ($0^{\circ}$ service) satellites, we produce machine learning models to cover Africa, Europe and East Asia.
Finally, we rely on observations from the Himawari-8 and Himawari-9 Advanced Himawari Imager (AHI), we produce models to cover the rest of Asia and Oceania.
A summary of the data channels available on the satellites is given in \cref{section:channels-summary}.

\subsection{GPM Combined Radar-Radiometer Precipitation (GPM CORRA)}
The GPM CORRA product~\citep{gpm2b} combines observations from the precipitation radar and microwave imager on board the Tropical Rainfall Measuring Mission (TRMM) and GPM CO satellites to produce the most accurate and globally ($65 ^\circ$ N and S) available precipitation estimates. In this study, GPM CORRA v07 precipitation retrievals from the GPM CO satellite are used as ground truth for training our machine learning models.

\subsection{IMERG}
IMERG \citep{Huffman2020} is a global, half-hourly precipitation product with a $0.1 ^\circ \times 0.1 ^\circ$ spatial resolution that combines retrievals from PMW and longwave IR 
observations. Corrections based on monthly precipitating gauge data are also applied to the precipitation estimates.
IMERG precipitation estimates are considered to be one of the most robust satellite-based products available~\citep{PRADHAN2022112754}. The IMERG algorithm generates precipitation estimates at three different latencies to serve various needs.
IMERG Early, available 4 hours after observation, provides near real-time estimates for time-critical applications. The second, IMERG Late, available after 14 hours, is more complete and accurate. The final product, IMERG Final, is released 3.5 months later with the inclusion of monthly rain gauge data. IMERG Final serves as the research-grade product for scientific analysis.
In this study, we benchmark against IMERG Early because of its intended use for low latency applications and IMERG Final because of its superior quality. In all cases we use the seventh version of the product (v07B).
In \Cref{section:transfer-learning}, we introduce a pre-training scheme that also makes use of the IMERG-Final precipitation retrievals.

\subsection{PDIR-Now}
The PERSIANN family of global precipitation products are one of the most widely used IR-based precipitation estimates available.
PDIR-Now~\citep{PDIR-Now}, a half-hourly retrieval with a $0.04 ^\circ \times 0.04 ^\circ$ spatial resolution is the latest and most advanced in this line of products making it an important baseline for evaluating the performance of our models.
While PDIR-Now relies solely on the longwave IR channel from GEO satellites as input, its framework for establishing the Tb-R relationship is a multi-step process.
The initial Tb-R relationships are calibrated using the IMERG Merged PMW precipitation. Subsequently, these pre-calibrated curves are dynamically adjusted using precipitation climatology data, specifically, WorldClim version 2~\citep{essd-7-275-2015} over land and PERSIANN-CDR over water.

\subsection{Convective Rain Rate (CRR)}
The Convective Rain Rate (CRR) is rainfall retrieval produced by EUMETSAT NWCSAF that provides estimates at a $3km \times 3km$ spatial resolution for regions covered by the Meteosat GEO satellites.
Owing to work done by the GCRF Swift project~\citep{TheAfricanSWIFTProjectGrowingScienceCapabilitytoBringaboutaRevolutioninWeatherPrediction, africanowcastwithnwcsaf}, it has gained popularity in operational use among a number of NMS in Africa, in addition to weather agencies that have already been using it operationally in the Europe. This makes it a key benchmark dataset against which to compare. The CRR algorithm uses the VIS ($0.6 \mu m$), longwave IR, and WV ($6.2 \mu m$) channels to produce a precipitation retrieval.

\section{Methodology}

\subsection{Training dataset}

\begin{figure*}[t]
  \centering
%   \hspace{15mm} % Manually adjust horizontal position.
  \includegraphics[width=0.9\linewidth]{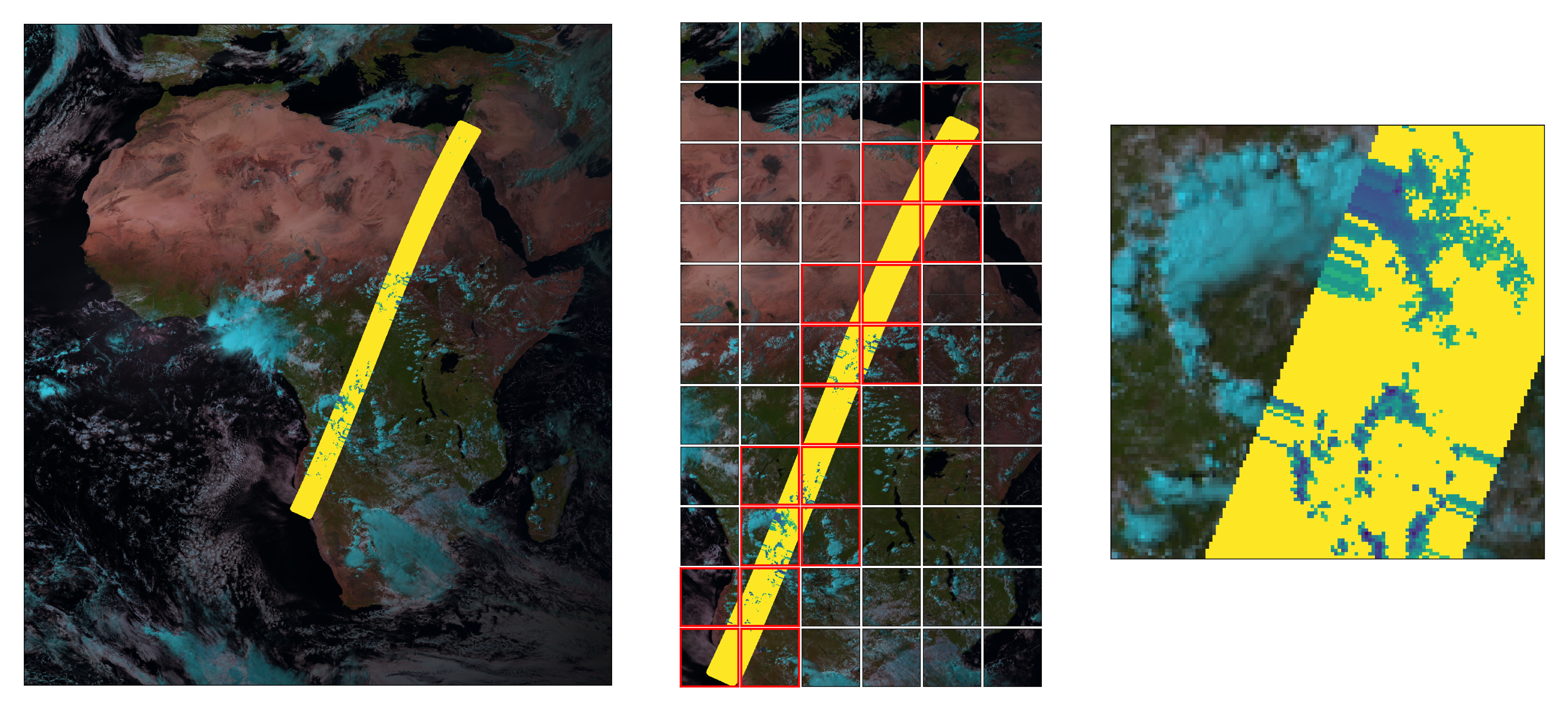}
  \caption{Example creation pipeline. (Left) shows a false color image of the Meteosat $0^{\circ}$ observation for April, 9 2022, 13:15 UTC, overlaid with the GPM CORRA precipitation swath between the start and end time of the MSG snapshot. (Middle) The patches produced by the example creation pipeline. Valid patches, that is, those that have corresponding GPM CORRA observations are highlighted in red. (Right) An example input-output pair showing the Meteosat $0^{\circ}$ observations (false color image) overlaid with the corresponding GPM CORRA precipitation retrieval.}
  \label{fig:example_creation}
\end{figure*}

The Oya precipitation estimation model is an example of a \emph{supervised learning} problem. That is, the first step is to produce a set of \emph{training examples} where input data are paired with correct output. In our case, the training examples for Oya are generated by collocating the GEO satellite observations with the GPM CORRA precipitation estimates. This is done separately for each series of GEO satellites.
Let $t_{start}$ and $t_{end}$ be the start and end times for a GEO satellite observation. We fetch the GPM CORRA scans that are within the time window from $t_{start}$ to $t_{end}$, giving us a pair of collocated GEO and precipitation observations.
The pair of examples are mapped onto a global equirectangular grid with a nominal spatial resolution of $5km$ at the equator.
%which is the native resolution of CORRA.
Lastly, the collocated examples are divided into square patches as shown in  \Cref{fig:example_creation} where only the patches with a valid GPM CORRA swath (shown in red) are retained. The patches have a grid size of $128 \times 128$ covering an area of $640 km \times 640 km$. The experimental results underlying this choice is presented in ~\Cref{section:patch-size}.

CORRA observations are available from 1998, however, we are limited by the availability of the current generation of GEO satellites. Training data for the Meteosat IODC service and the GOES satellites are drawn from between 2016 to 2021 for training on GPM CORRA. For the Himawari satellite, we fetch the training data from 2017 to 2021 and for the Meteosat $0^{\circ}$ degree service, the training data is drawn from 2014 to 2021. Since GPM CORRA observations have limited spatial sampling, this limits the amount of training data, which can lead to slow training and model overfitting, where the model does not generalize well to  observations that were not in the training set. To address this, we also construct a \emph{pretraining} dataset using IMERG Final as ground truth.
The IMERG Final data for pretraining with the Meteosat $0^{\circ}$ satellite is drawn from 2004 to 2021, giving us nearly two decades of historical data for pretraining the Oya models. Pretraining data for the other GEO satellites are fetched from the same time period as was used for the training on the GPM CORRA observations..
In all cases, observations from 2022 are set aside and curated into a validation set for evaluating the performance of the models. 

More formally, let $X \equiv \mathbb{R}^{H \times W \times C}$ be the space of GEO observations, and let $Y \equiv \mathbb{R}^{H \times W}$ be the space of GPM CORRA-v07 observations. In general, a GPM CORRA-v07 observation is not available for every point in the $640 km \times 640 km$ patch, so we define $M \equiv \mathbb{B}^{H \times W}$ to be the space of masks that specify the grid locations where a GPM CORRA-v07 observation exists (1 if there is a valid observation, 0 otherwise).

The supervised learning problem takes a set of $N$ training examples $T \equiv \{ (\vx_1, \vy_1, \vm_1), \ldots, (\vx_N, \vy_N, \vm_N) \} \subseteq X \times Y \times M$ to learn a parameterized function $f_{\thetamin}: X \rightarrow Y$ that minimizes a continuous loss function $g: M \times Y \times Y \rightarrow \mathbb{R}$. That is,

\begin{equation}
    \thetamin = \mathop{\hbox{\rm arg min}}_{\theta} \sum_{(\vx, \vy, \vm) \in T} g(\vm, \vy, f_{\theta}(\vx)).
\label{eq:training-objective}
\end{equation}

In practice, we implement $f$ as a neural network, where $\theta$ are the parameters of the network, and $g$ is the $L_2$ loss.

\begin{equation}
    g(\vm, \vy, \vy') = \sum_{ij} 
\begin{cases}
    (\vy_{ij}' - \vy_{ij})^2 & \text{if $\vm_{ij} = 1$} \\
    0 & \text{otherwise}
\end{cases}
\label{eq:loss-fn}
\end{equation}

\subsection{Handling Class imbalance}\label{section:handling-imbalance}

\begin{figure*}[t]
\small
  \centering
  \begin{tabular}{cc}
       \includegraphics[width=0.45\linewidth]{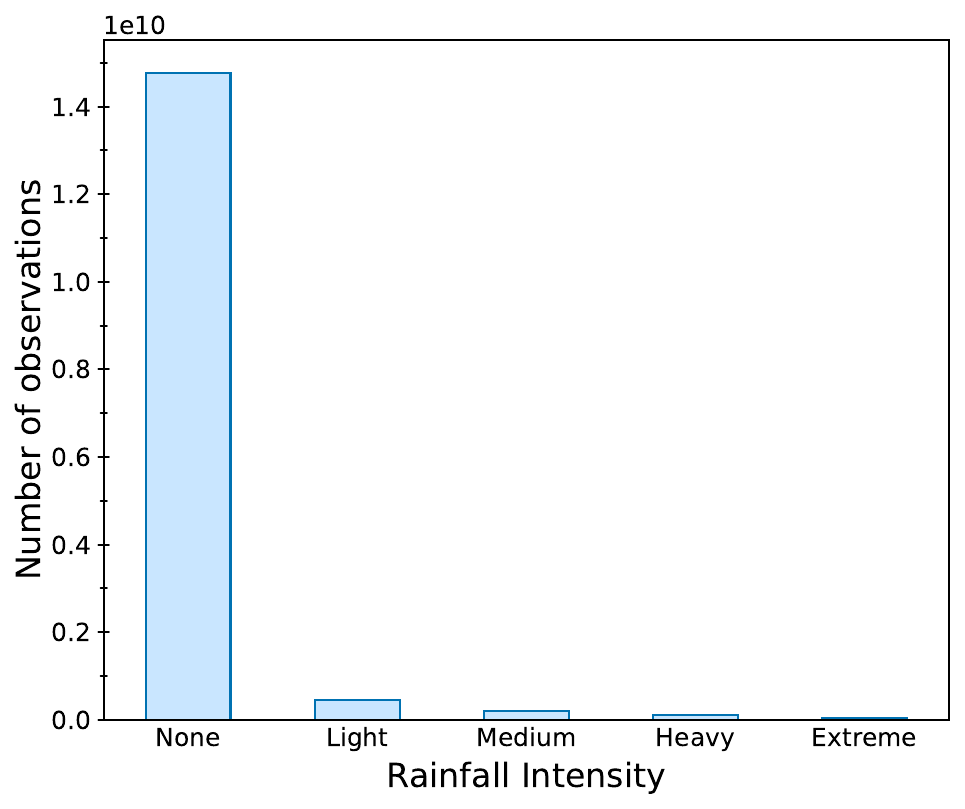} & 
       \includegraphics[width=0.45\linewidth]{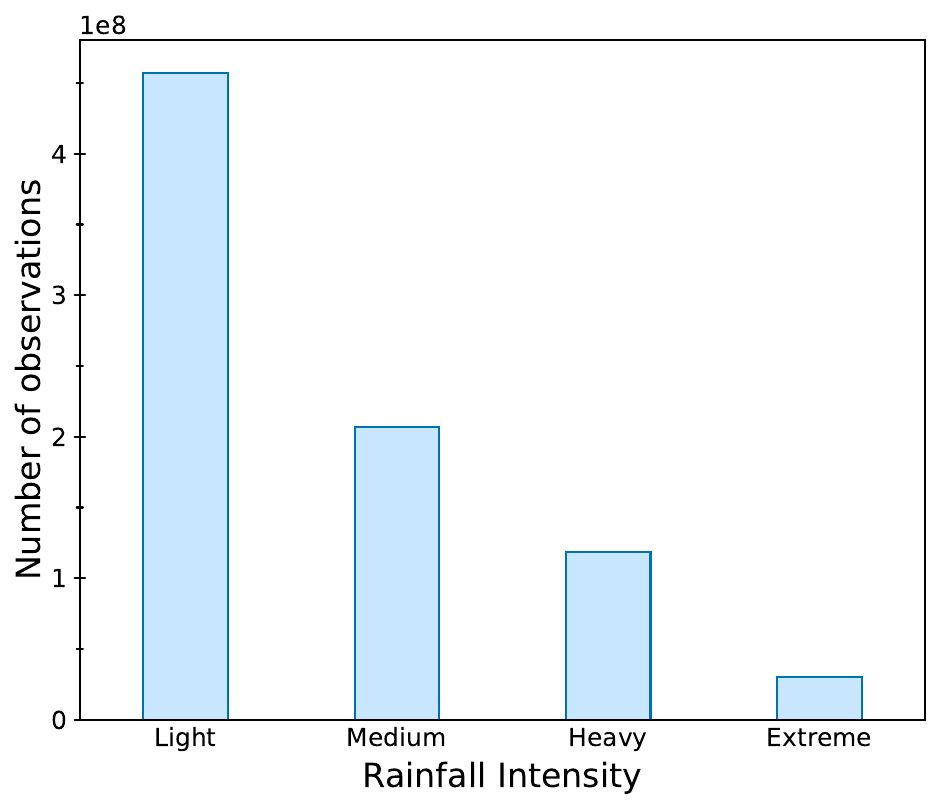} \\
       a & b
  \end{tabular}
  \caption{Distribution of GPM CORRA v07 observations. (a) Distribution of no-precipitation, light, medium, heavy and extreme precipitation events, showing the imbalance between the no-precipitation and precipitation observations. (b) Distribution of precipitation events from (a) showing that in the absence of no-precipitation observations, light precipitation observations heavily outweigh the medium to extreme precipitation events.}
  \label{fig:class_imbalance}
\end{figure*}

\Cref{fig:class_imbalance} shows a plot of the distribution of the GPM CORRA precipitation observations accumulated across the full extent of the GPM CO satellite from 2014 to 2022.
The following thresholds are used for grouping the observations: 0.2 $mm\:h^{-1}$ (light precipitation), 1.0 $mm\:h^{-1}$ (medium precipitation), 2.4 $mm\:h^{-1}$ (heavy precipitation), 7.0 $mm\:h^{-1}$ (extreme precipitation) (closely matching levels in the AMS Glossary of Meteorology~\citep{ams-glossary-rain}).

An acute imbalance is immediately obvious between the no-precipitation and precipitation observations with the former outweighing the latter by a factor of 18 to 1.
This makes it difficult to train accurate models as the model is heavily biased in favour of the no-precipitation observations while our testing criterion places more emphasis on the precipitation observations~\citep{cao2019learningimbalanceddatasetslabeldistributionaware}.
Common approaches for long-tailed recognition like loss-re-weighting  and data re-sampling struggle in these cases since the re-sampling/re-weighting ratios have to be quite large to compensate for the imbalance. This introduces a significant distribution shift between the training and test sets. 

We address this problem by decomposing the estimation task into two distinct tasks: a precipitation/no-precipitation classification task and a quantitative precipitation estimation task as has been done in some previous retrieval algorithms ~\citep{AScreeningMethodologyforPassiveMicrowavePrecipitationRetrievalAlgorithms}.
We train two models, one a classifier to detect precipitation events (without estimating the amount of precipitation) and the other a regression model to estimate the rate of precipitation.
The regression model is only trained on precipitation events and is not applied to no-precipitation events. Thus, the regression model does not have to deal with the imbalance between the no-precipitation events and the precipitation observations in~\Cref{fig:class_imbalance}(a).
The outputs of the two models are multiplied to produce the final precipitation estimate. With this, $f(\vtheta, \vx)$ becomes:
\begin{equation}
    f(\vtheta, \vx) = f_1(\theta_1, \vx) \cdot f_2(\theta_2, \vx).
\label{eq:two-step-model}
\end{equation}

This approach only partially addresses the imbalance problem. Although not as severe as in \Cref{fig:class_imbalance}(a), the distribution of precipitation and no-precipitation events in the precipitation/no-precipitation model remains skewed in favour of the no-precipitation class. 
However, since $f_1(\theta_1, \vx)$ is a classification model, we are able to leverage traditional long tailed recognition approaches to help with the imbalance. Specifically, we utilize label weights in the loss function to increase the performance of the model on the precipitation class.
Furthermore, when there is precipitation, the intensity tends to follow a power law ~\citep{powerlaw, WhyDoPrecipitationIntensitiesTendtoFollowGammaDistributions} which also needs to be compensated for in the regression model (\Cref{fig:class_imbalance}(b)).
To do this, we train the regression model to predict the logarithm of the amount of precipitation which reduces the skew of the data, making the distribution of precipitation more symmetric. The log transformation is reversed before metrics are computed.
Existing techniques for learning from imbalanced data focus on targets with categorical indices~\citep{yang2021delvingdeepimbalancedregression}.
\cite{yang2021delvingdeepimbalancedregression} recently introduced Label Distribution Smoothing (LDS) as a technique for imbalanced regression. LDS uses kernel density estimation to smooth the empirical label distribution, creating an effective label density that accounts for the overlap in information between data points with similar target values. 
We implement a cost-sensitive re-weighting scheme where the loss function of the regression model is re-weighted with the inverse of the LDS estimated label density for each target.
\begin{figure*}[t]
\small
  \centering
  \begin{tabular}{cc}
       \includegraphics[width=0.3\linewidth]{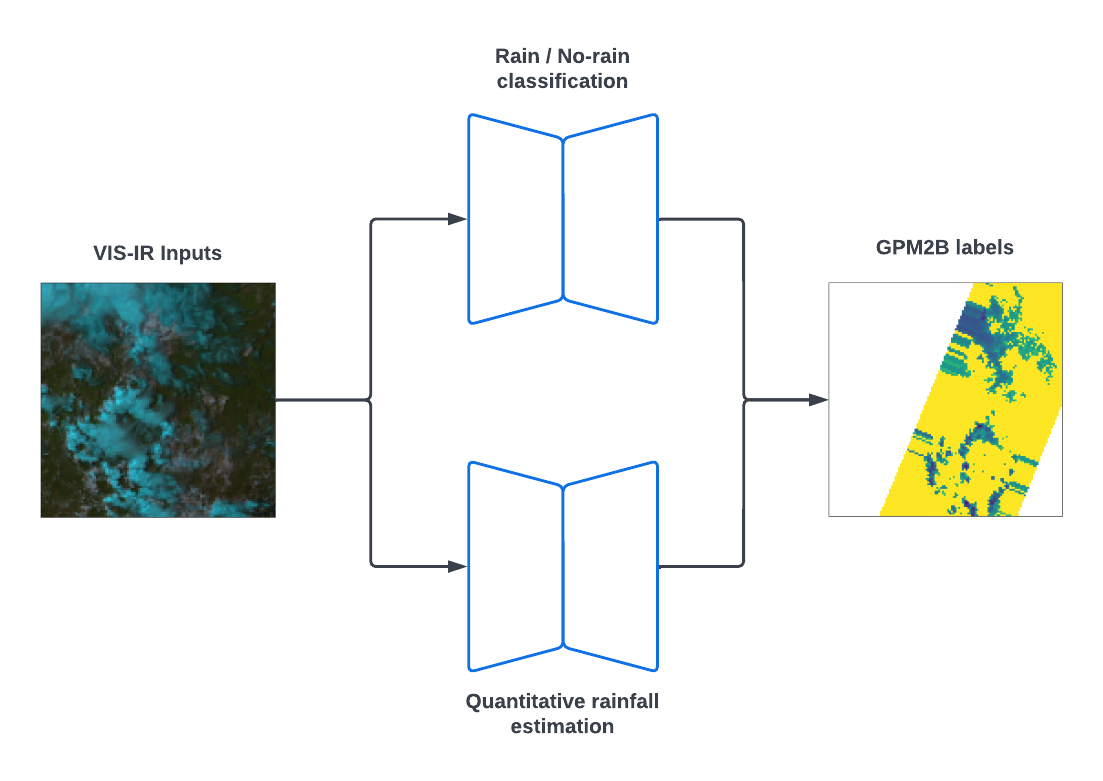} & 
       \includegraphics[width=0.6\linewidth]{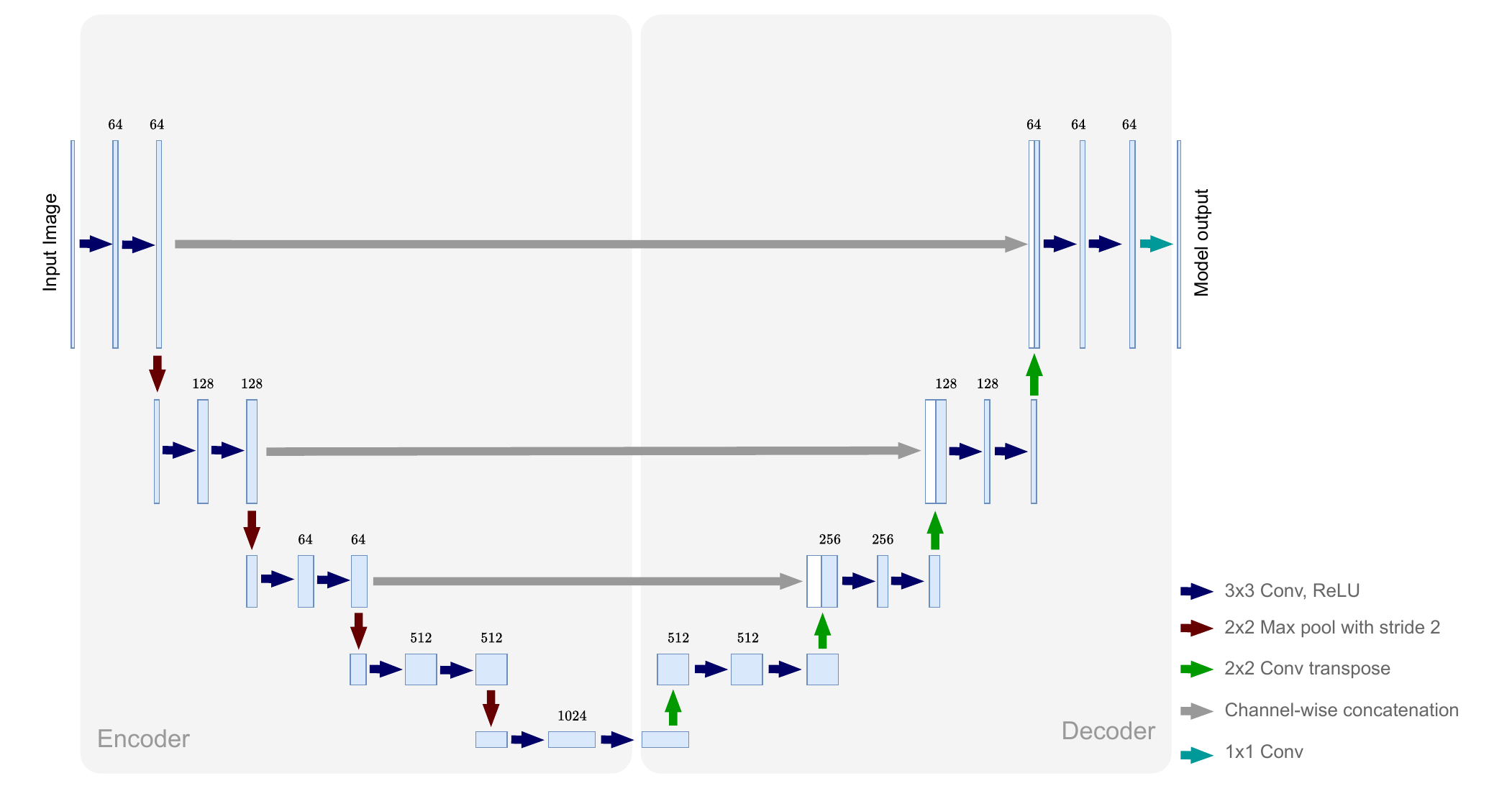} \\
       a & b
  \end{tabular}
  \caption{(a) The Oya precipitation estimation model consisting of two UNet models: a classification model to detect precipitation events, and a regression model to estimate the actual amount of precipitation. (b) the UNet architecture used for each model, which consists of an encoder and a decoder with skip connections.}
  \label{fig:architecture}
\end{figure*}

\subsection{Data Augmentation}

A primary challenge with using GPM CORRA estimates as ground truth for training a model is the sparsity of its observations. As has been already mentioned, the learning signal for a given [GEO, GPM CORRA] pair comes from a small subset of the number of pixels in the target GPM CORRA observations.
Thus, even with a large training set, the effective number of training samples is much smaller. Deep learning models are known to be data hungry~\citep{LeCun2015-en, Goodfellow-et-al-2016}, with this inherent need for vast amounts of data stemming from the complexity of the underlying architectures and the model's reliance on learning intricate patterns from the data available~\citep{bengio2014representationlearningreviewnew}.
Sparse observations make the model prone to overfitting. We make use of \emph{data augmentation} to alleviate this problem. 

Data augmentation is a common technique in machine learning used to artificially increase the size and diversity of the training dataset by creating modified copies of existing data~\citep{yang2022image}. This process helps models generalize better and reduces overfitting by exposing them to a wider range of variations ~\citep{wong2016understanding, cirecsan2010deep}. Care must however be taken when choosing augmentation methods for physical data like satellite imagery, as transformations should ideally preserve the physical relationship between the input (GEO channels) and the output (precipitation). Therefore, we restrict our augmentations to geometric transformations that maintain the spatial integrity and physical plausibility of the scenes. Specifically, during training, we randomly apply horizontal flips, vertical flips, and $90^{\circ}$ rotations to the input GEO patches and their corresponding precipitation targets.

\subsection{Transfer Learning}\label{section:transfer-learning}

Transfer learning is key ingredient of many state-of-the-art deep learning approaches where the model is pre-trained on a different but related problem with a large dataset where overfitting is less likely.  Once pre-trained, the model is fine-tuned for the target problem by training it on a possibly much smaller target dataset~\citep{transferfeatures}.
Transfer learning leads to better performance compared to training from scratch, especially when the target dataset is small~\citep{huh2016makesimagenetgoodtransfer, Oquab2014LearningAT}.

In our case, we pretrain our models on a related precipitation estimation task using IMERG Final v07B estimates as the ground-truth, and then fine-tune using GPM CORRA precipitation estimation as the target.
Although IMERG Final is a lower quality source compared to GPM CORRA, it is evenly gridded with full spatial coverage and a long historical record, and is well-suited for training machine learning models. In comparison, the target GPM CORRA dataset is spatially very sparse, but as we show below it is more than sufficient for fine tuning the model.
The lower accuracy of IMERG Final is well tolerated in this case. As well, IMERG uses CORRA for inter-satellite calibration, so its histogram of precipitation rates should be strongly related to that of CORRA.
As \cite{noisylabels} show, deep neural networks are capable of generalizing from noisy training data and demonstrated high test performance after training on data for which the true labels were massively outnumbered by incorrect labels.

\subsection{Neural network model}

As shown in \Cref{fig:architecture}, Oya consists of two models whose outputs are combined to produce the final precipitation estimate. The architecture in each case is a UNet~\citep{unet}, a Convolutional Neural Network that has proven to be effective for various dense prediction tasks including segmentation of satellite images due to its ability to precisely localize regions of interest~\citep{openbuildings, Wang_2022, app14093712}. It consists of an encoder and a decoder with skip connections from the feature maps in the encoder to the decoder to enable precise localization.

The classification model is trained using the softmax cross entropy loss to identify precipitating pixels in the input image. The regression model on the other hand is trained to predict the log  of the amount of precipitation at each precipitating pixel with a mean squared loss function. Both models are trained using the AdamW~\citep{adamw} optimizer with a constant learning rate schedule. 

\begin{figure}
\begin{tabular}{cc}
    % \small
    % \centering
    $\vcenter{\hbox{\includegraphics[width=.4\linewidth]{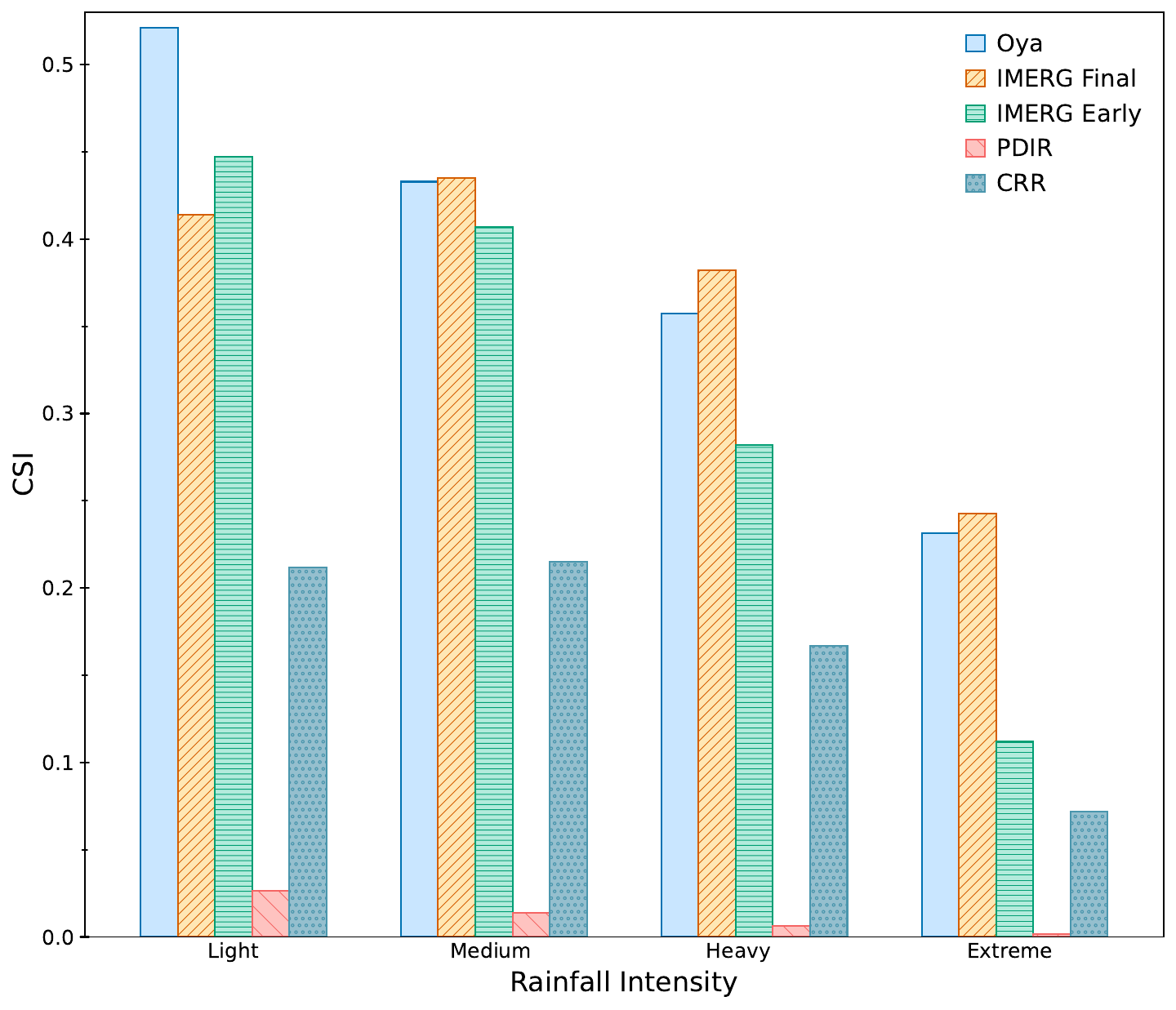}}}$ &
    % \small
    % \centering

        \setlength{\tabcolsep}{5pt}
        \begin{tabular}{lcccc}
        \toprule
            Retrieval    &  CSI  & POD & FAR & Bias  \\        
        \midrule
            CRR & 0.212 & 0.279 & 0.531 & 0.595 \\
            PDIR-Now & 0.026 & 0.075 & 0.961 & 1.935 \\
            IMERG Early & 0.447 & 0.686 & 0.438 & 1.221 \\
            IMERG Final & 0.414 & \textbf{0.746} & 0.518 & 1.550 \\
            Oya (Africa) & \textbf{0.521} & 0.705 & \textbf{0.334} & \textbf{1.058} \\

        \bottomrule
        \end{tabular} 
\\
a & b
\end{tabular}
\caption{Retrieval accuracy metrics of model trained over Africa against GPM CORRA observations in 2022. (a) Critical Success Index (CSI) for Oya, IMERG Final, IMERG Early, PDIR and CRR retrievals at different precipitation intensities. (b) CSI, Probability of Detection (POD), False Alarm Ratio (FAR) and Bias for each retrieval at a threshold of 0.2$mm\:h^{-1}$. The best result for each metric is in bold font.
\label{fig:africa_csi}
}
\end{figure}

\section{Results}

This section presents evaluation results of the Oya precipitation retrievals and is organized as follows. We first present the results of a model trained over the African subregion, followed by the results of extending this model to cover the entire Meteosat $0^{\circ}$ region.
Third, the results are presented for the models trained on each of the remaining GEO satellites to achieve quasi-global coverage. An analysis of Oya's accuracy globally is then presented.
Evaluation results are all computed against GPM CORRA v07 precipitation observations at a $5\times5$ km spatial resolution.

\textbf{Metrics}

We evaluate the performance of Oya using several standard categorical metrics for validating weather forecasts \citep{Stanski1989-uu}. These include the critical success index (CSI), probability of detection (POD), frequency bias index (Bias) and false alarm ratio (FAR).
Definition of these metrics are given in \cref{section:metric-definitions}.
% Given a precipitation threshold $T$, these metrics compute the ratio of pixels correctly classified or misclassified as being greater than or equal to $T$ to those that were not \citep{CreatingSyntheticRadarImageryUsingConvolutionalNeuralNetworks}.
% The following thresholds are used for the evaluations presented in this paper: 0.2 $\mathop{\ms{mm}} h^{-1}$ (light precipitation), 1.0 $\mathop{\ms{mm}}h^{-1}$ (medium precipitation), 2.4 $\mathop{\ms{mm}}h^{-1}$ (heavy precipitation), 7.0 $\mathop{\ms{mm}}h^{-1}$ (extreme precipitation) (closely matching levels in the AMS Glossary of Meteorology~\citep{ams-glossary-rain}).

\subsection{Africa} \label{africa_results}

\Cref{fig:africa_csi} presents the performance of Oya trained over Africa. Of the products that are presented, only CRR and PDIR-Now are based solely on GEO observations. As discussed in \Cref{data}, PDIR-Now is based solely on the longwave IR
% ($10.7 \mu m$)
channel while CRR uses observations from two other channels in addition to the longwave IR
% ($10.7 \mu m$) 
channel.
As such, they are the closest to an ``apples to apples'' comparison. IMERG Early and IMERG Final (v07B) rely primarily on higher quality PMW observations.
Oya significantly outperforms both CRR and PDIR-Now on all the precipitation thresholds. Compared to CRR, the CSI of the Oya retrievals are better by more than 30 points for light precipitation, 20 points for medium and heavy precipitation and 10 points for extreme precipitation. 
Oya estimates are also better than IMERG Early retrievals for all the precipitation thresholds. This result is noteworthy, considering that for the instances that are used in the evaluation, the IMERG algorithm has access to the full constellation of PMW sensors.

Furthermore, Oya retrievals are also comparable to IMERG Final precipitation estimates. Oya is better on low intensity precipitation events whiles IMERG Final is better for high intensity events. Thus with only GEO observations, we are able to produce precipitation estimates in real time that are comparable to a research-quality product that is produced after 3 months.

\Cref{fig:africa_csi} also presents results on the other metrics for a threshold of 0.2 $mm\:h^{-1}$. We find that with the exception of POD where IMERG Final is better, Oya's estimates are better on all the other metrics by large margins. For example, the FAR of Oya retrievals are lower by more than 20 points compared to the FAR of IMERG Final. 

\subsection{Results for Meteosat 0$^\circ$}\label{section:all-meteosat}

\begin{figure}
\begin{tabular}{cc}
    % \small
    % \centering
    $\vcenter{\hbox{\includegraphics[width=.4\linewidth]{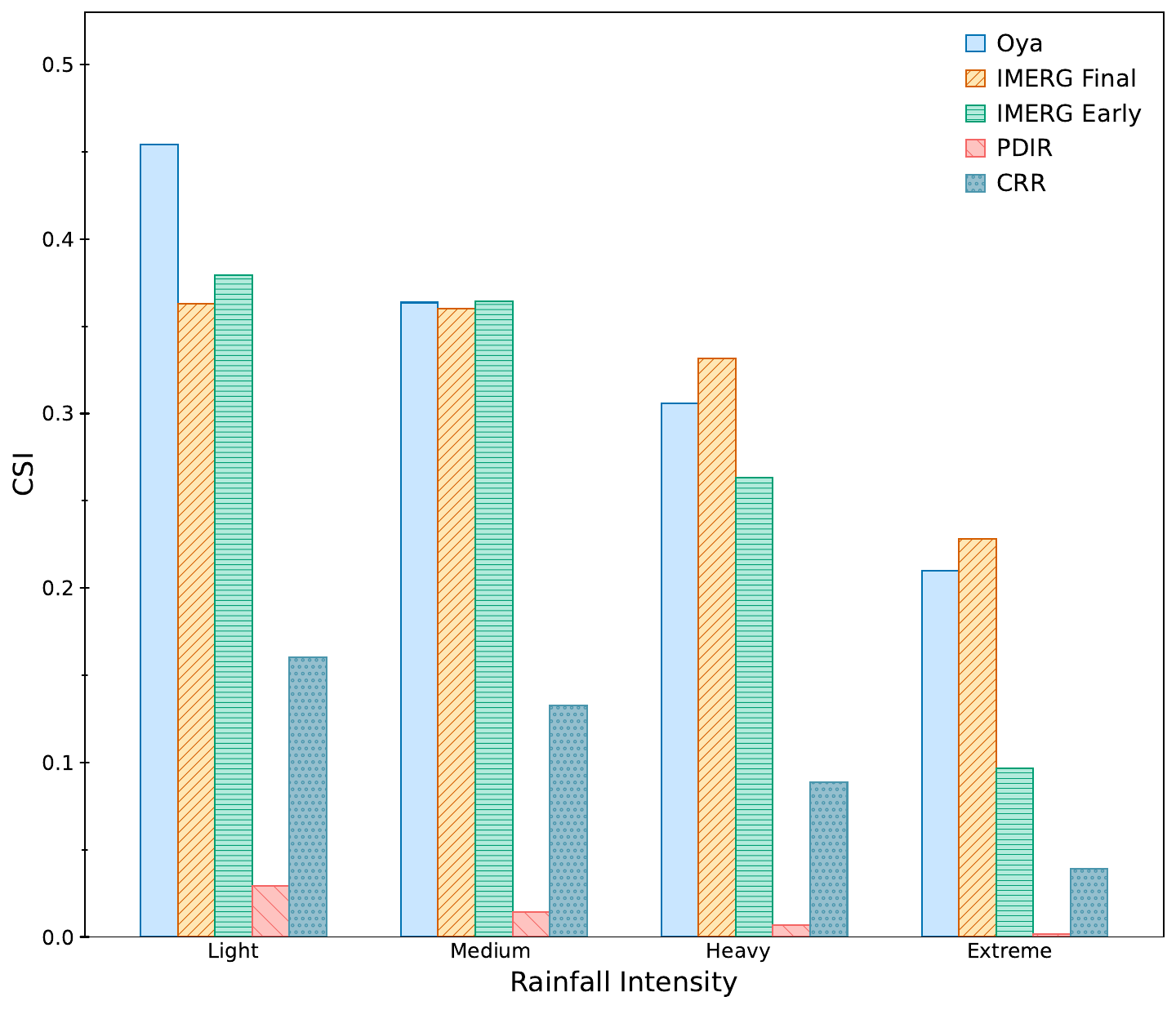}}}$ &
    % \small
    % \centering

        \setlength{\tabcolsep}{5pt}
        \begin{tabular}{lcccc}
        \toprule
            Retrieval    &  CSI  & POD & FAR & Bias  \\        
        \midrule
            CRR & 0.160 & 0.227 & 0.649 & 0.647 \\
            PDIR-Now & 0.029 & 0.062 & 0.947 & 1.168 \\
            IMERG Early & 0.379 & 0.610 & 0.499 & 1.217 \\
            IMERG Final & 0.363 & \textbf{0.666} & 0.556 & 1.500 \\
            Oya (Meteosat 0$^{\circ}$) & \textbf{0.454} & 0.638 & \textbf{0.388} & \textbf{1.043} \\

        \bottomrule
        \end{tabular} 
\\
a & b
\end{tabular}
\caption{As in \Cref{fig:africa_csi} but for the model trained over Meteosat 0$^{\circ}$ coverage. \label{fig:meteosat0_csi}}
\end{figure}

\Cref{fig:meteosat0_csi} presents the results of extending the model in \Cref{africa_results} to cover the entire Meteosat 0$^{\circ}$ region.
Oya retrievals continue to outperform the other baselines namely CRR, PDIR-Now and IMERG Early while remaining competitive with IMERG Final. However, we also observe a uniform decline in performance compared to the model trained only over Africa.
There are several factors that can explain this performance drop. One is limb darkening~\citep{nasa-limb-darkening}, where the temperature distribution of thermal radiation deviates at higher latitudes because the slant path through the atmosphere at higher zenith angles is longer than for equatorial observations, which tends to shift the weighting profile of the channels to higher (and colder) altitudes. 
This effect is nonlinear, causing intermediate temperatures to appear significantly cooler as the slant angle increases.
Another likely cause is parallax shift, which causes misplacement of high elevation cloudtops. This parallax shift is progressively more pronounced in the higher latitudes. We experimented with additional inputs that could help the model deal with the parallax effect.
These include cloud top height, latitude and latitude coordinates, azimuth and zenith angles  and elevation. Yet, the models trained with these additional inputs did not perform better than the models trained only with GEO inputs. More work is thus needed to compensate for these effects.

\subsection{Extending to other geosynchronous satellites}

\begin{figure}
\begin{tabular}{cc}
    % \small
    % \centering
    $\vcenter{\hbox{\includegraphics[width=.4\linewidth]{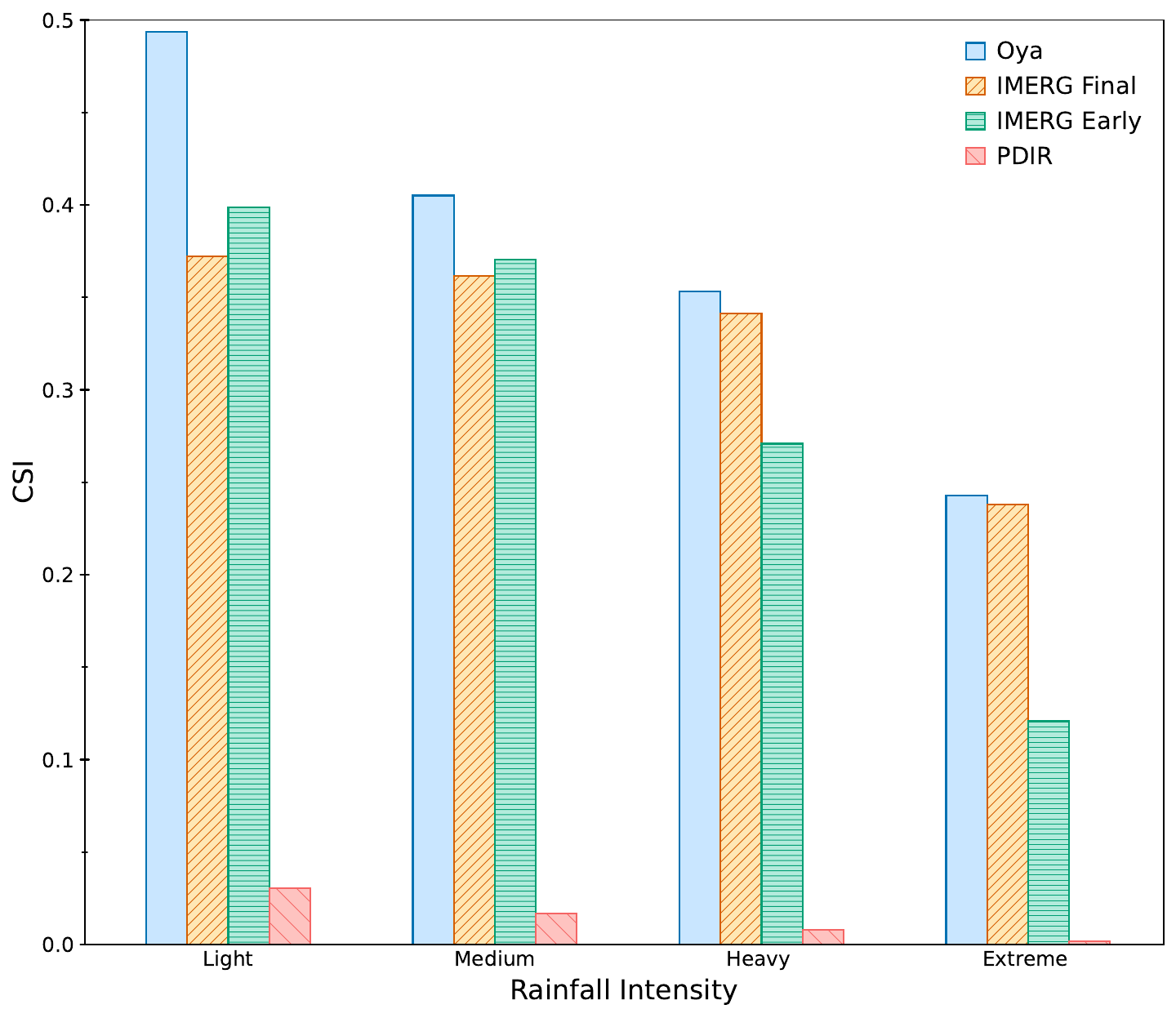}}}$ &
    % \small
    % \centering

        \setlength{\tabcolsep}{5pt}
        \begin{tabular}{lcccc}
        \toprule
            Retrieval    &  CSI  & POD & FAR & Bias  \\        
        \midrule
            PDIR-Now & 0.031 & 0.052 & 0.930 & 0.741 \\
            IMERG Early & 0.399 & 0.647 & 0.491 & 1.270 \\
            IMERG Final & 0.372 & \textbf{0.696} & 0.556 & 1.567 \\
            Oya (Himawari) & \textbf{0.494} & 0.673 & \textbf{0.350} & \textbf{1.035} \\

        \bottomrule
        \end{tabular}
    \\
    a & b

\end{tabular}
\caption{As in \Cref{fig:africa_csi} but for the model trained over Himawari coverage. \label{fig:himawari_csi}
}
\end{figure}

\begin{figure}
\begin{tabular}{cc}
    % \small
    % \centering
    $\vcenter{\hbox{\includegraphics[width=.4\linewidth]{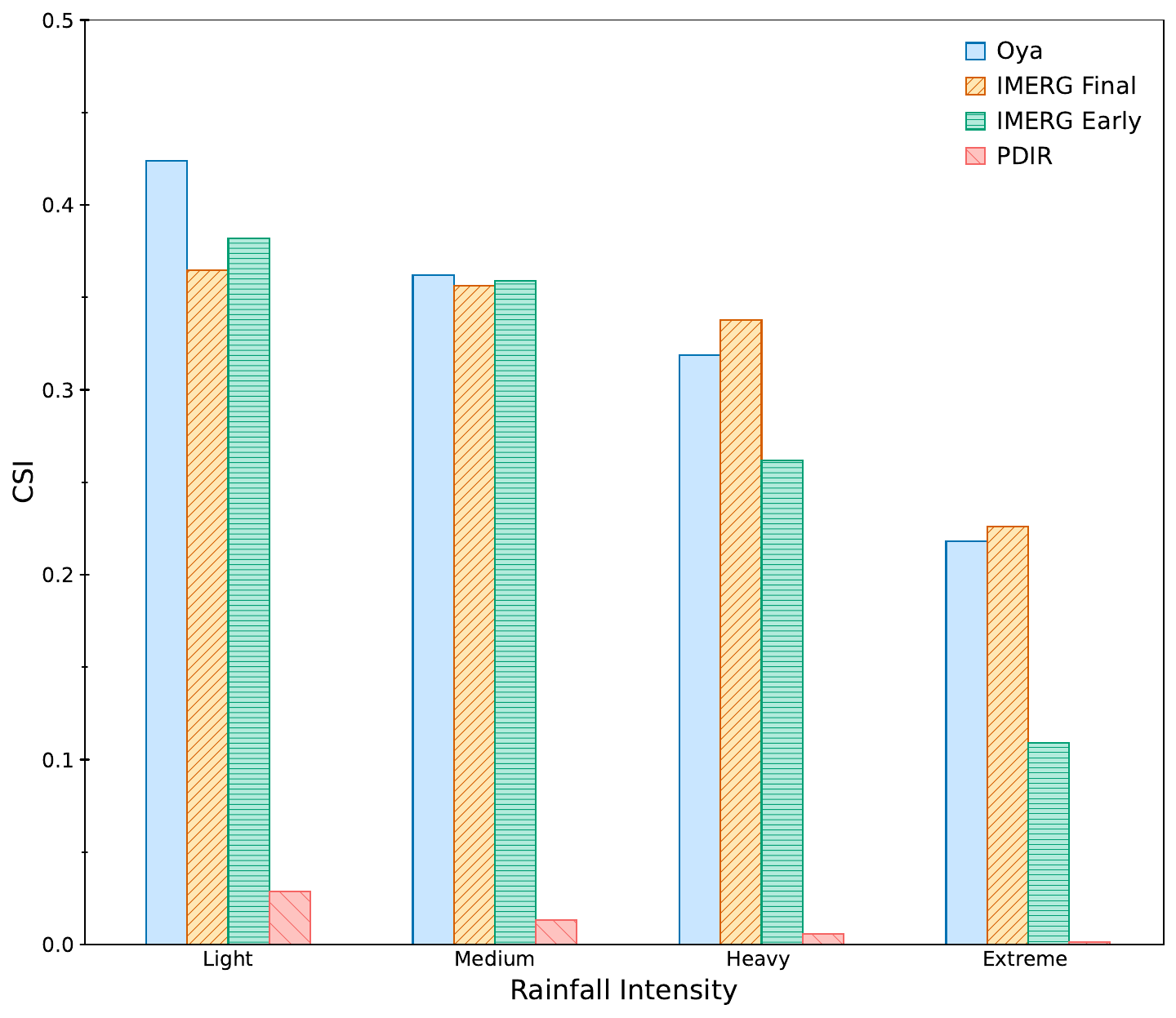}}}$ &
    % \small
    % \centering

        \setlength{\tabcolsep}{5pt}
        \begin{tabular}{lcccc}
        \toprule
            Retrieval    &  CSI  & POD & FAR & Bias  \\        
        \midrule
            PDIR-Now & 0.029 & 0.057 & 0.946 & 1.052 \\
            IMERG Early & 0.382 & 0.618 & 0.500 & 1.237 \\
            IMERG Final & 0.365 & \textbf{0.672} & 0.556 & 1.516 \\
            Oya (Meteosat 45$^{\circ}$) & \textbf{0.424} & 0.597 & \textbf{0.406} & \textbf{1.005} \\

        \bottomrule
        \end{tabular}
    \\ 
    a & b

\end{tabular}
\caption{As in \Cref{fig:africa_csi} but for the model trained over Meteosat 45$^{\circ}$ coverage. \label{fig:meteosat45_csi}
}
\end{figure}

To achieve global coverage, we separately train models for each of the remaining GEO satellite sources, namely Himawari, Meteosat 45$^{\circ}$, and GOES-16. The historical catalogue of the GOES-18 satellite begins in 2022 which is also the time period used for validation, so no models were trained on the GOES-18 satellite. 
Rather, since the GOES-16 and GOES-18 satellites share the same specifications, the model trained on the GOES-16 satellite observations was used for the GOES-18 region.

For Himawari, we see in \Cref{fig:himawari_csi} that the CSI of Oya retrievals is higher than that of IMERG Final for all the precipitation thresholds. However the performance trend over the precipitation thresholds remain similar to that of the Meteosat 0$^{\circ}$ models. 
Oya estimates are much better than the IMERG Final (which is the best of the comparison products presented in this study) for light amounts of precipitation and comparable for higher intensity precipitation events.
On the other metrics, Oya estimates are better on all except the POD. The FAR of Oya estimates for example is lower by 20 percentage points compared to that of IMERG Final. 

\begin{figure}
\begin{tabular}{cc}
    % \small
    % \centering
    $\vcenter{\hbox{\includegraphics[width=.4\linewidth]{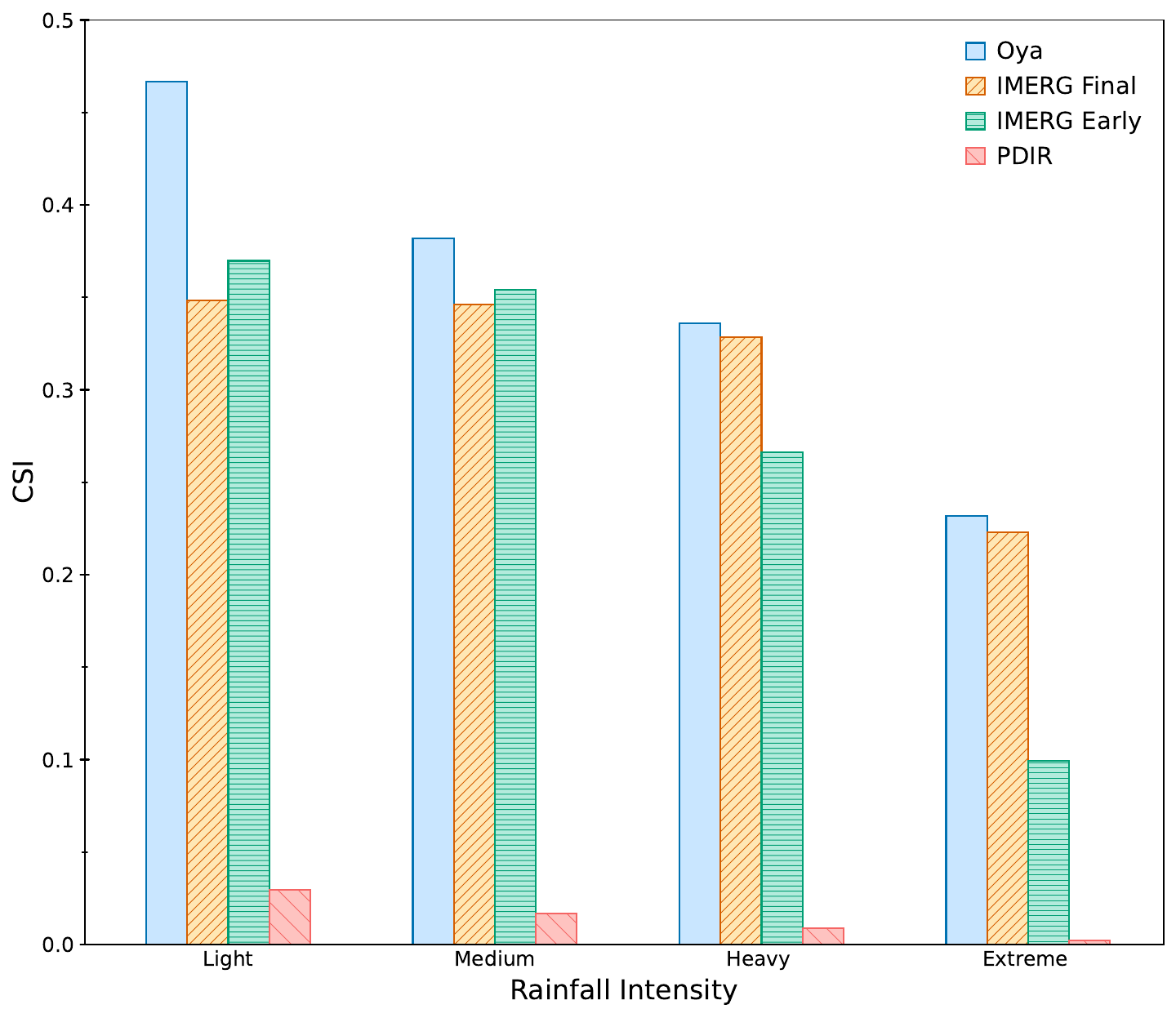}}}$ &
    % \small
    % \centering

        \setlength{\tabcolsep}{5pt}
        \begin{tabular}{lcccc}
        \toprule
            Retrieval    &  CSI  & POD & FAR & Bias  \\        
        \midrule
            PDIR-Now & 0.030 & 0.056 & 0.941 & 0.959 \\
            IMERG Early & 0.370 & 0.598 & 0.508 & 1.215 \\
            IMERG Final & 0.349 & \textbf{0.651} & 0.571 & 1.520 \\
            Oya (GOES-16) & \textbf{0.467} & 0.637 & \textbf{0.365} & \textbf{1.002} \\

        \bottomrule
        \end{tabular}
    \\ 
    a & b

\end{tabular}
\caption{As in \Cref{fig:africa_csi} but for the model trained over GOES-16 coverage. \label{fig:goes16_csi}
%Retrieval accuracy metrics of model trained over GOES-16 coverage. (a) Critical Success Index (CSI) for Oya, IMERG Final, IMERG Early and PDIR-Now retrievals at different precipitation intensities. (b) CSI, Probability of Detection (POD), False Alarm Ratio (FAR) and Bias for each retrieval at a threshold of 0.2$mmh^{-1}$. The best result for each metric is in bold font.
}
\end{figure}

Over the Meteosat 45$^{\circ}$ and GOES-16 region (\Cref{fig:meteosat45_csi}, \Cref{fig:goes16_csi}), Oya continues to show stong performance compared to the baselines. The CSI is better than IMERG Early and PDIR-Now while remaining competitive with IMERG Final.
As seen in the previous results (\Cref{fig:africa_csi}, \Cref{fig:meteosat0_csi}, \Cref{fig:himawari_csi}) Oya is also better on all the other metrics with the exception of the POD.

\subsection{Global Analysis of Retrieval Accuracy}

\begin{figure*}[t]
  \centering
%   \hspace{15mm} % Manually adjust horizontal position.
  \includegraphics[width=0.9\linewidth]{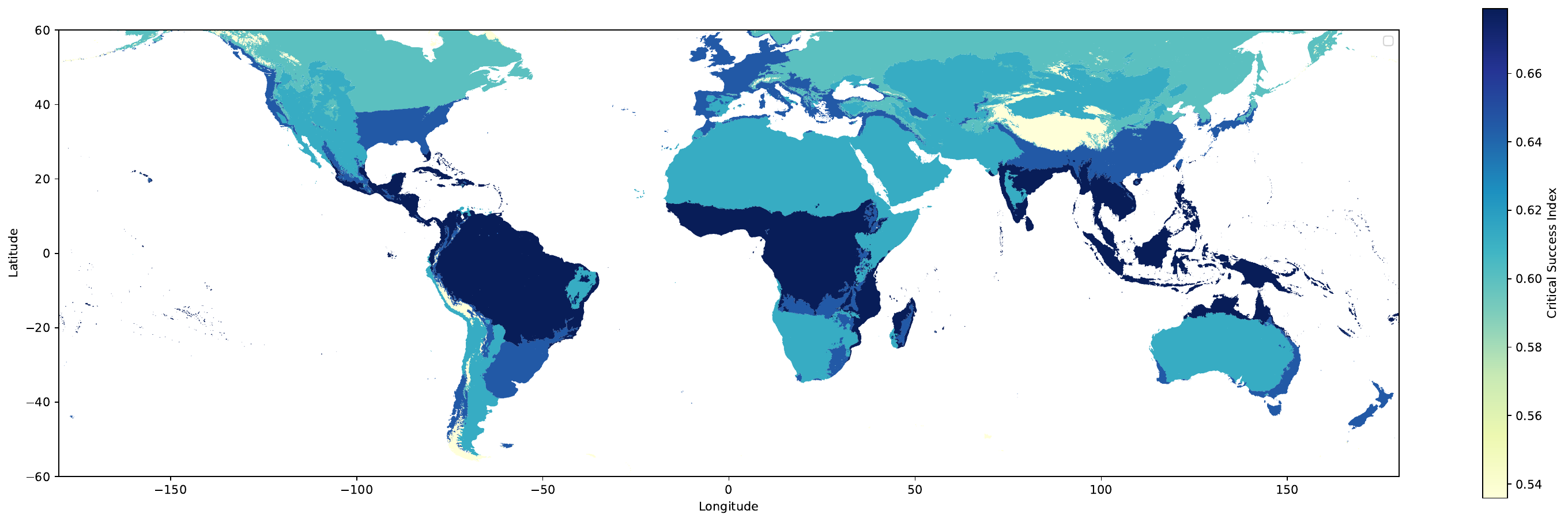}
  \caption{Global map of CSI for Oya precipitation estimates agains GPM CORRA observations in 2022}
  \label{fig:global_csi_map}
\end{figure*}

To provide a more detailed understanding of the model's performance in relation to different climatic and geographic features, we analyzed the spatial distribution of Oya's retrieval accuracy.
\Cref{fig:global_csi_map} presents a global map of the CSI for the Oya retrievals, evaluated against GPM CORRA observations at a 0.2 $mm\:h^{-1}$ threshold.
%The analysis is performed for five climate zones namely the tropical, arid, temperate, cold and polar regions. 

The results show that Oya's retrievals are most accurate in the tropics and decline in performance with latitude, consistent with the observation in \Cref{section:all-meteosat}. In addition, performance is poor in dry, high altitude regions like the Tibetan plateau. More study is needed to understand this performance drop. Between these two regions, the temperate zone is the next most accurate followed in descending order by the arid and cold zones. 

\subsection{Case Study}

\begin{figure*}[h]

  \centering

  \begin{tabular}{cccc}
        (a) Meteosat & (b) CORRA & (c) Oya & (d) IMERG Final \\

       \includegraphics[width=0.22\linewidth]{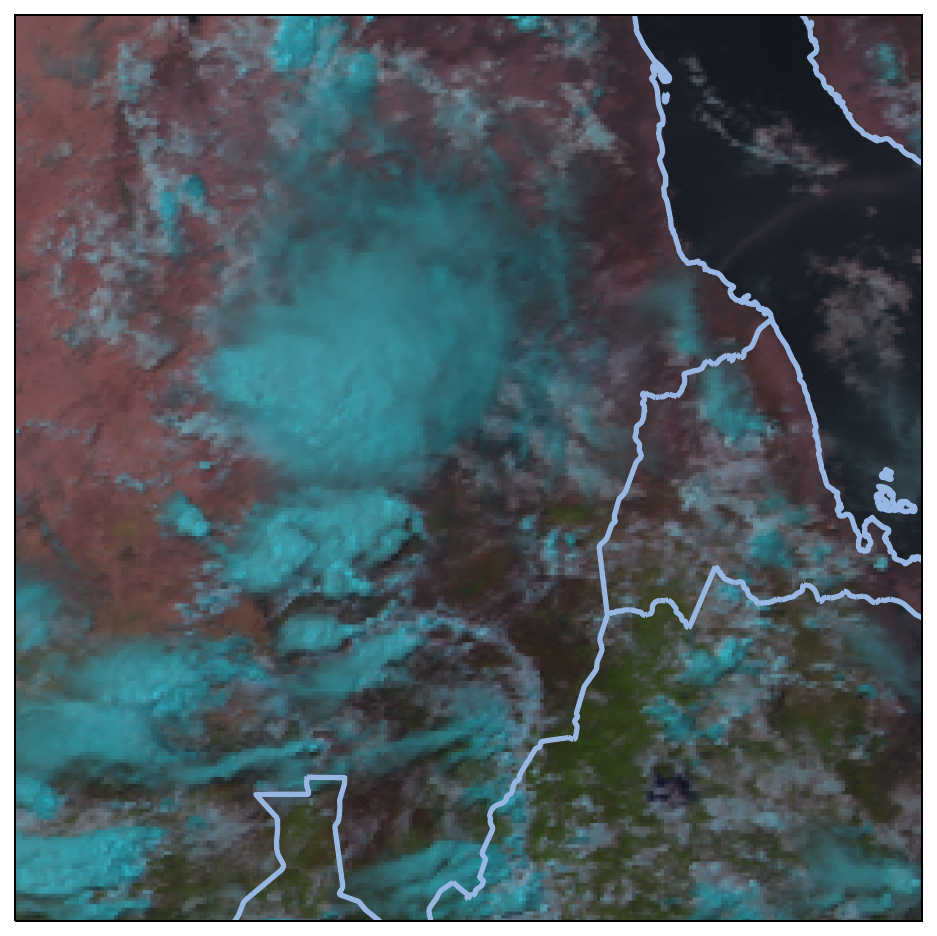} & 
       \includegraphics[width=0.22\linewidth]{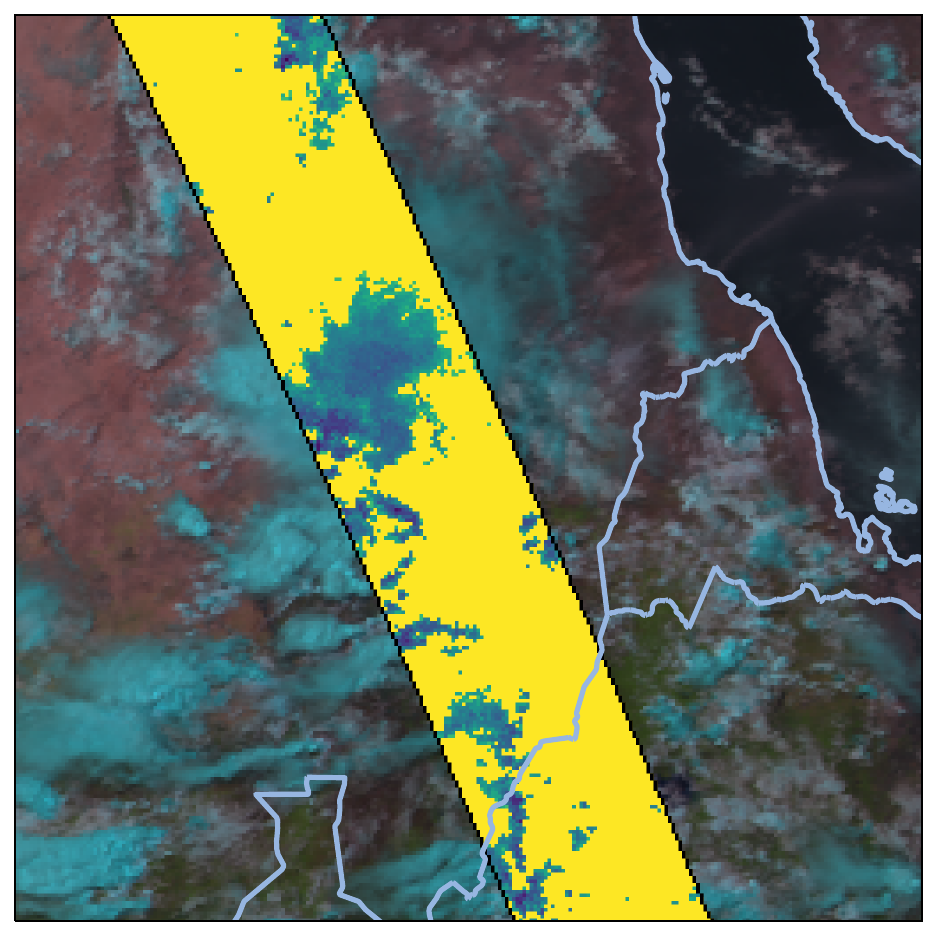} &
       \includegraphics[width=0.22\linewidth]{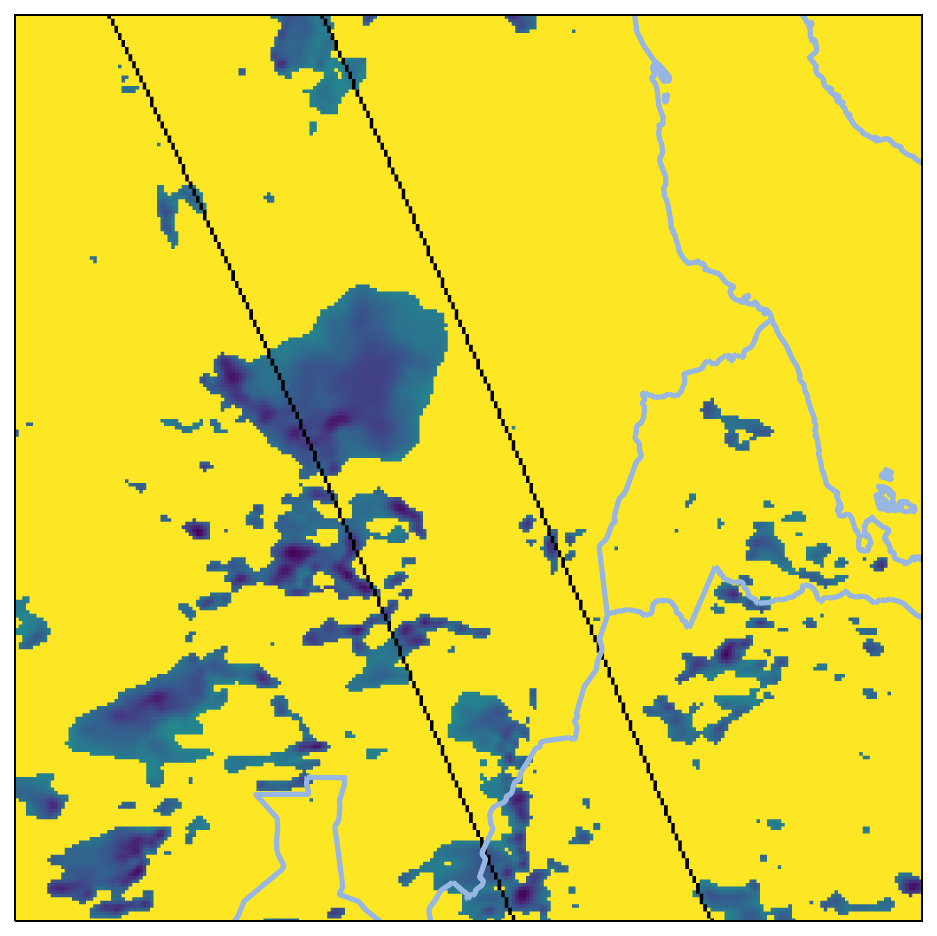} &
       \includegraphics[width=0.22\linewidth]{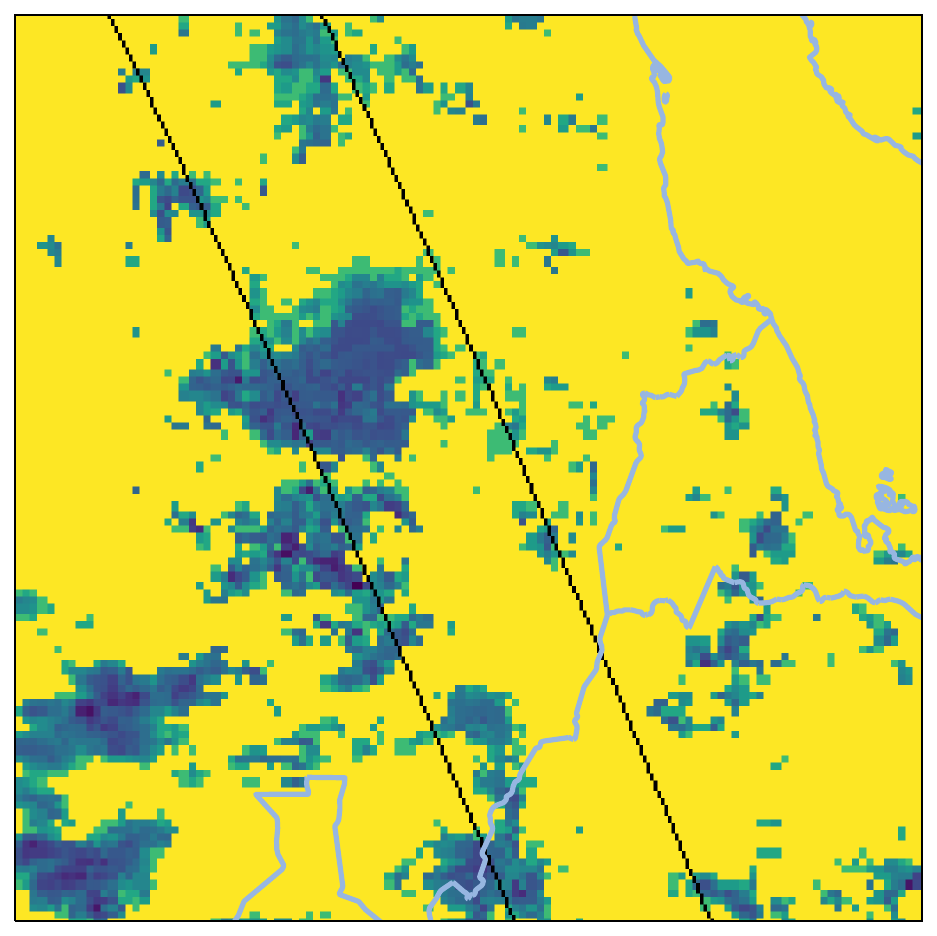} \\
       
       (e) IMERG Early & (f) PDIR-Now & (g) CRR &  \\
       
       \includegraphics[width=0.22\linewidth]{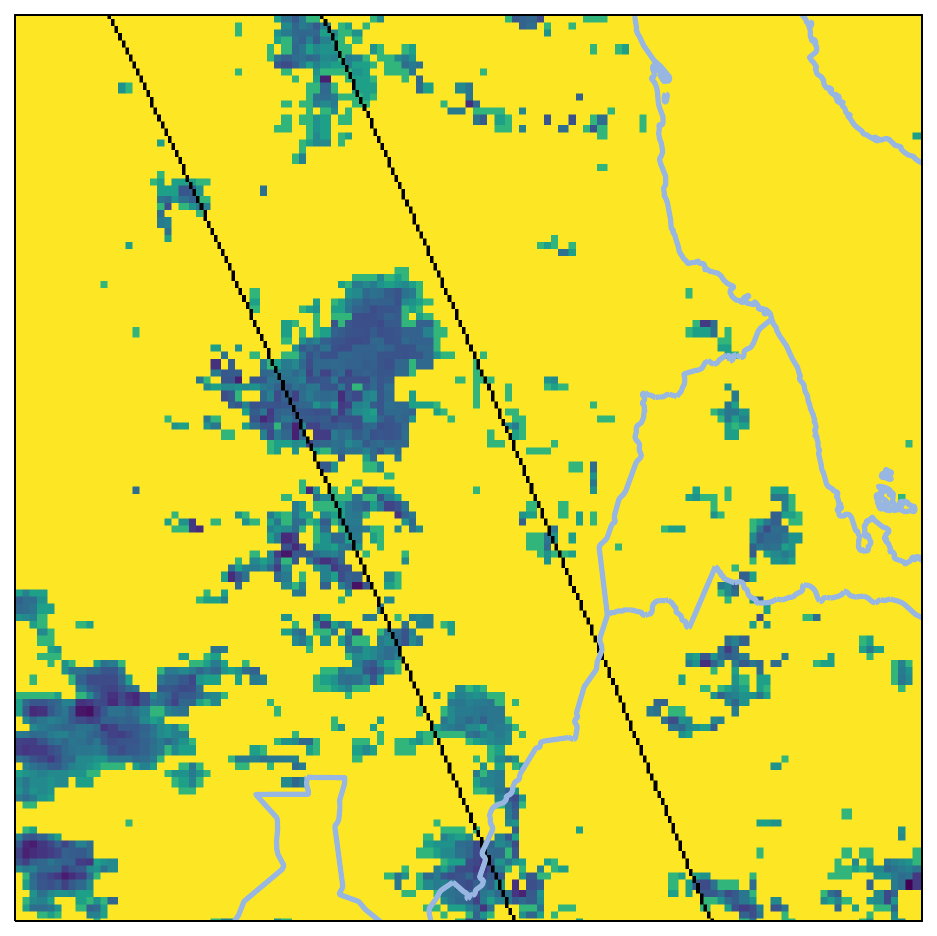} & 
       \includegraphics[width=0.22\linewidth]{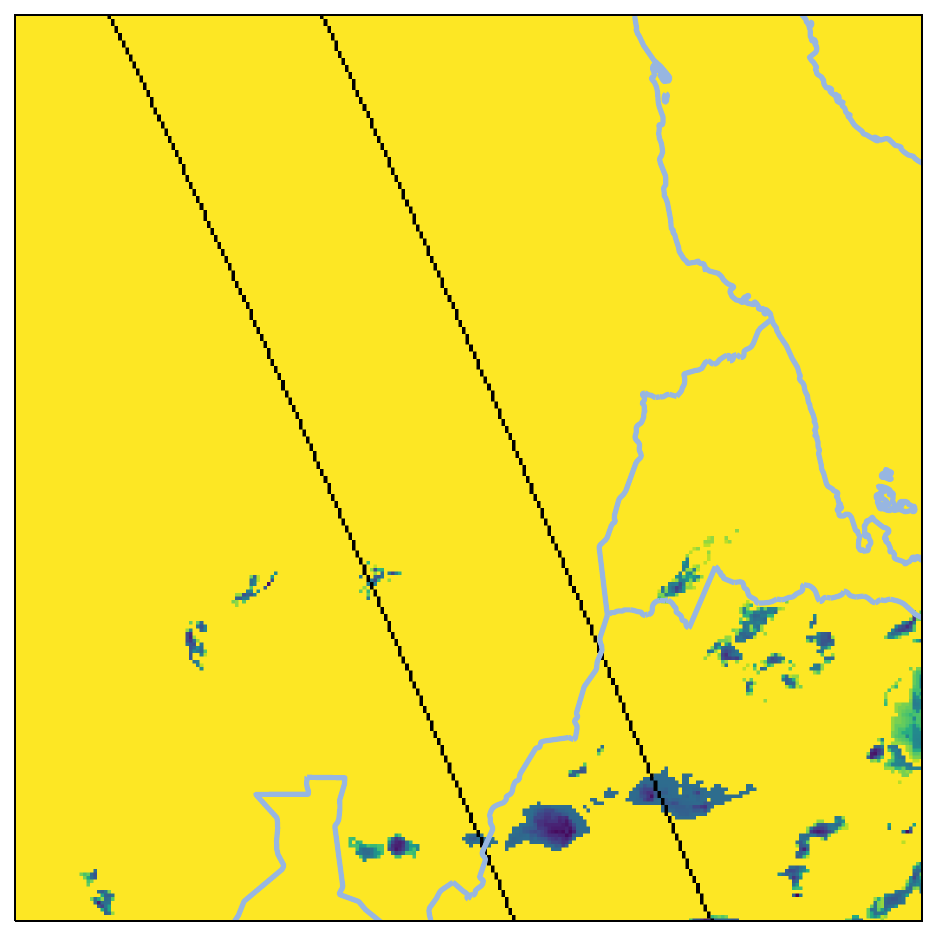} &
       \includegraphics[width=0.22\linewidth]{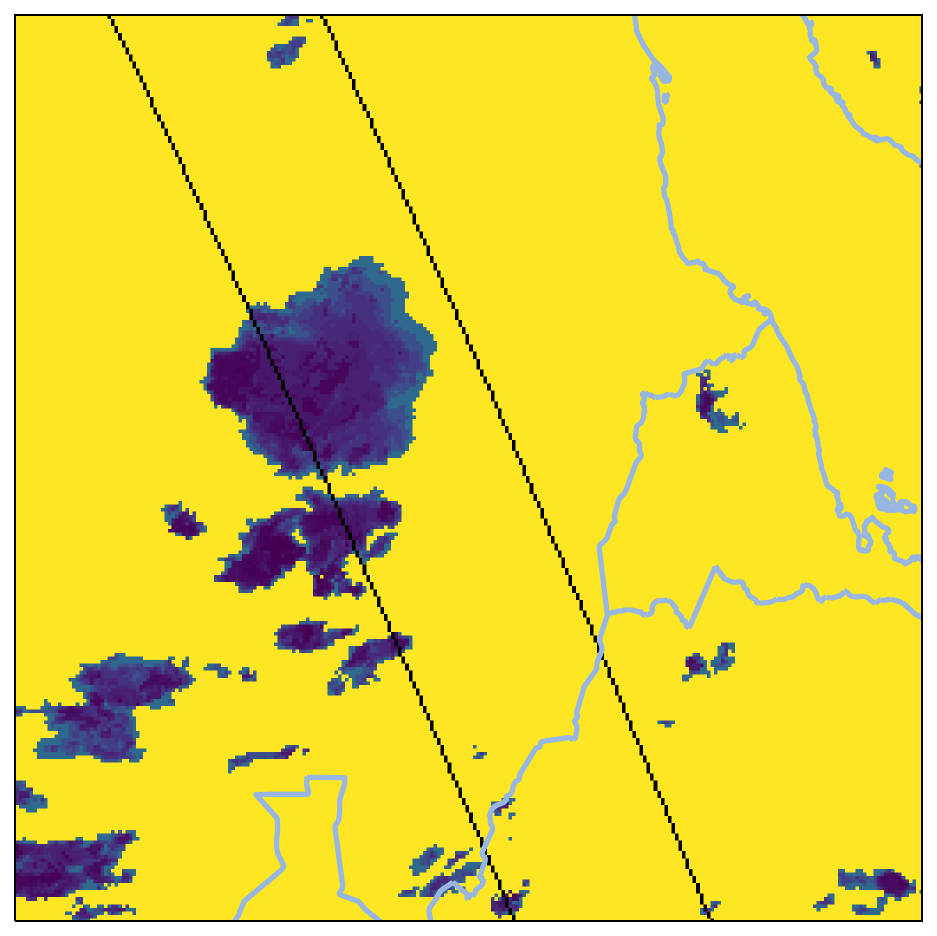} & \\
       
  \end{tabular}
  \caption{GPM overpass for a storm over South Sudan on August 8, 2022, at 12:45:00 UTC. (a) Meteosat false color image. Precipitation retrievals from (b) GPM CORRA swath overlaid on the Meteosat false color image, (c) Oya, (d) IMERG Final, (e) IMERG Early, (f) PDIR-Now, and (g) CRR. The black dotted lines indicate the GPM CORRA swath while country boundaries are shown in blue.}
  \label{fig:case_study_liberia_estimates}
\end{figure*}

\begin{figure}
\begin{tabular}{cc}
    % \small
    % \centering
    $\vcenter{\hbox{\includegraphics[width=.5\linewidth]{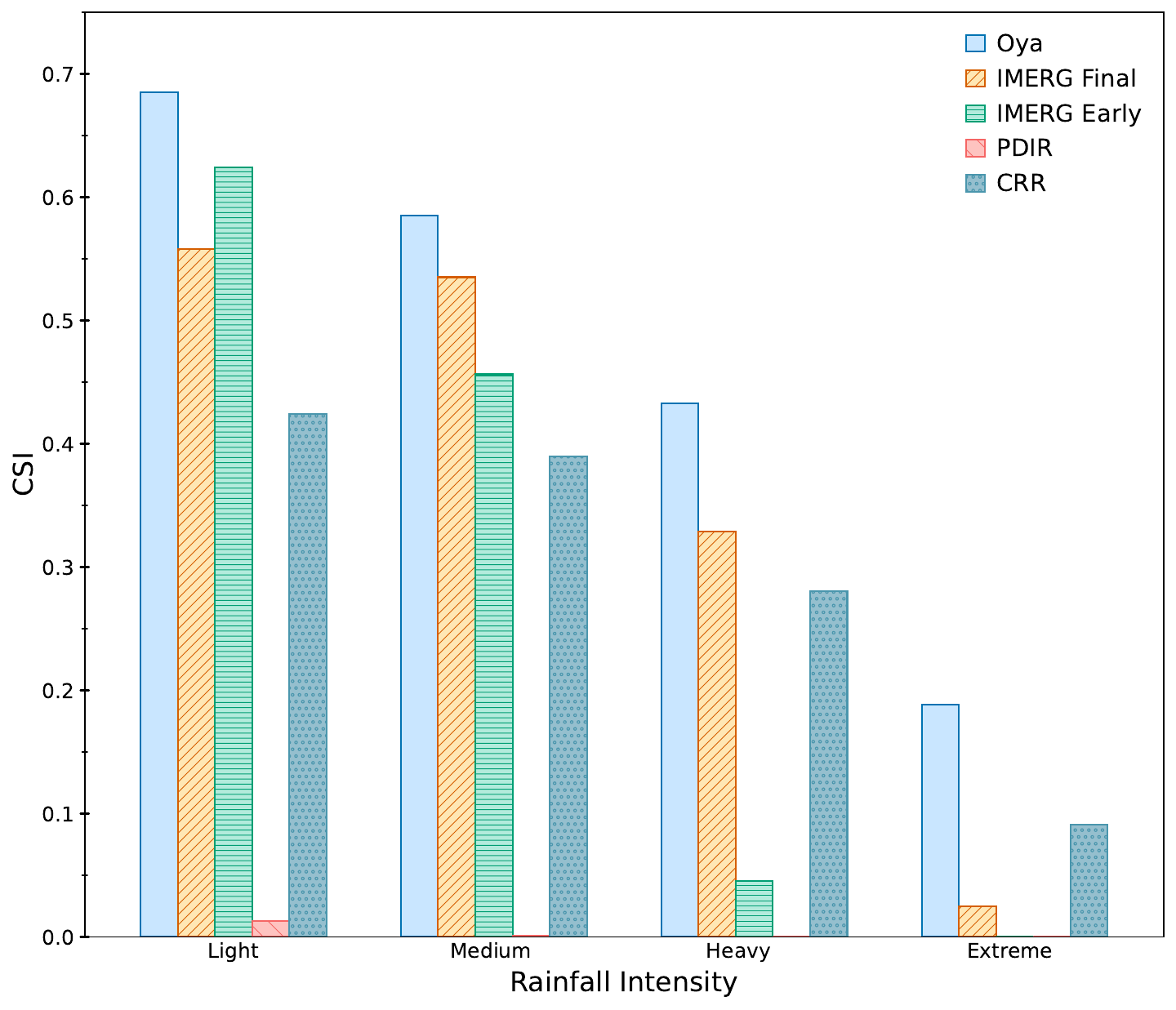}}}$ &
    % \small
    % \centering

        \setlength{\tabcolsep}{5pt}
        \begin{tabular}{lcccc}
        \toprule
            Retrieval    &  CSI  & POD & FAR & Bias  \\        
        \midrule
            CRR & 0.424 & 0.535 & 0.329 & 0.797 \\
            PDIR-Now & 0.013 & 0.014 & 0.869 & 0.105 \\
            IMERG Early & 0.624 & 0.742 & \textbf{0.203} & \textbf{0.931} \\
            IMERG Final & 0.558 & 0.804 & 0.354 & 1.244 \\
            Oya & \textbf{0.685} & \textbf{0.867} & 0.234 & 1.132 \\

        \bottomrule
        \end{tabular}
    \\ 
    a & b

\end{tabular}
\caption{As in \Cref{fig:africa_csi} but for the storm shown in \Cref{fig:case_study_liberia_estimates}. \label{fig:case_study_liberia_csi}
%Retrieval accuracy metrics for storm shown in \Cref{fig:case_study_liberia_estimates}. (a) Critical Success Index (CSI) for Oya, IMERG Final, IMERG Early, PDIR-Now and CRR retrievals at different precipitation intensities. (b) CSI, Probability of Detection (POD), False Alarm Ratio (FAR) and Bias for each retrieval at a threshold of 0.2$mmh^{-1}$. The best result for each metric is in bold font.
}
\end{figure}

We consider the case of an overpass of GPM CORRA over a storm in South Sudan on August 8, 2022 at 12:45:00 UTC and compare the performance of Oya against the the other products.
\Cref{fig:case_study_liberia_estimates} displays a false colour image of the Meteosat observation alongside the precipitation retrieval from the GPM CO satellite, Oya, IMERG, PDIR-Now and CRR. The accuracy metrics with respect to the GPM CORRA data is shown in \Cref{fig:case_study_liberia_csi}.

The Oya estimate is revealed to be better than the CRR and  PDIR-Now retrieval across all precipitation thresholds. For example, compared to CRR, the CSI of the Oya estimate is better by almost 20 percentage points for light, medium and heavy precipitation and by more than 10 percentage points on extreme precipitation. This trend holds for the other metrics as well in \cref{fig:case_study_liberia_estimates}.

Compared to IMERG Early and IMERG Final, the CSI of the Oya estimate is also higher on all precipitation thresholds. On light precipitation, Oya is better than IMERG Final by more than 10 percentage points. The FAR of the IMERG Early retrieval is however better than that of the Oya estimate. A look at the Bias scores shows that this is because the IMERG Early retrieval underestimates rain leading to a bias less than one while the Oya retrieval overestimates rain leading to a Bias greater than one.

\section{Ablations}

\emph{Ablations} are methods used on machine learning to measure the relative importance of design choices as well as input data. In our case we perform ablation studies to that measure the performance impact of various design choice in the Oya precipitation estimation setup. For simplicity, the models in this section are all trained over Africa.

\begin{table}[h]
    \small
    \centering
    \begin{tabular}{llcccc}
    \toprule
                  &             & \multicolumn{4}{c}{CSI}   \\ 
        Ablation & Description &  Light & Medium & Heavy & Extreme  \\ 
    \midrule
        GEO Channels & With $10.7 \mu m$ Channel & 0.319 & 0.260 & 0.219 & 0.145 \\
                     & All Channels & 0.490 & 0.401 & 0.323 & 0.205 \\
                     
    \bottomrule
        Data Augmentation & Without Augmentation & 0.435 & 0.371 & 0.264 & 0.117 \\
                    & With Augmentation & 0.490 & 0.406 & 0.305 & 0.148 \\
    
    \bottomrule
        Pretraining & Without Pretraining & 0.490 & 0.401 & 0.323 & 0.205 \\
                    & Pretrained on IMERG Final & 0.521 & 0.433 & 0.357 & 0.231 \\
    
    \bottomrule
        Patch Size & 32 & 0.472 & 0.378 & 0.297 & 0.180 \\
                   & 64 & 0.491 & 0.400 & 0.326 & 0.204 \\
                   & 128 & 0.494 & 0.390 & 0.325 & 0.209 \\
    
    \bottomrule
        LDS & Without LDS & 0.490 & 0.406 & 0.305 & 0.148 \\
            & With LDS & 0.490 & 0.401 & 0.323 & 0.205 \\

    \bottomrule
    
    \end{tabular}
    \caption{ \label{tab:ablations} Ablation studies of design choices in Oya precipitation estimation model.}
\end{table}

\subsection{Input Geostationary Channels}\label{section:number-of-channels}

In this work we utilize the full spectrum of the available GEO channels. However, many IR-based precipitation estimation products such as PDIR-Now and CRR make use of only a subset of the channels (VIS, longwave IR and WV for CRR and longwave IR in the case PDIR-Now).
We investigate the performance benefits of using all the channels as opposed to using only a subset in this section. Specifically, following the approach of PDIR-Now, we train a  model that only uses the longwave IR 
% ($10.7 \mu m$)
channel of the GEO imagery and compare the performance with that of a model trained with all the available bands in  \Cref{tab:ablations}.
The model trained with all of the channels outperforms the model trained with only the longwave IR channel by large margins. 
In terms of the CSI, the model that uses all channels performs better by more than 10 points for light, medium, and heavy precipitation, while its performance is better by more than 5 points for extreme precipitation.

\subsection{Impact of Data Augmentation}\label{section:data-augmentation-ablation}

Oya makes use of data augmentation during training to improve the model's generalization. Operations that are applied include flips (horizontal and vertical) and $90^{\circ}$ rotations. 

In this section we perform an ablation over this choice to ascertain its usefulness. We train a model without the use of data augmentation and compare that against the performance of another model trained with data augmentation in \Cref{tab:ablations}. The model trained with data augmentation outperforms the model trained without it in all precipitation classes. It is better by about 6 points on light precipitation and 3 points on the remaining classes demonstrating that data augmentation indeed improves the model's generalizability.

In \Cref{section:transfer-learning}, we presented a pre-training scheme where the models are first pretrained on IMERG Final and the weights transferred to the target GPM CORRA estimation task. We investigate the usefulness of the this pre-training step and report the downstream GPM CORRA estimation performance in \Cref{tab:ablations}. We find that pretraining on IMERG Final leads to consistent performance boosts across all precipitation thresholds compared to a model trained from scratch. 

Early experiments (not shown) also demonstrate carried out also show that this performance boost is independent of the pretraining dataset. Pretraining on TRMM also led to similar performance gains. Since the TRMM satellite is no longer operational we are not able to use it as our standard pretraining dataset. However, TRMM is a viable option training models for earlier GEO satellites. 

\subsection{Impact of the patch size}\label{section:patch-size}

In this section, we investigate the effect of the input patch size on the accuracy of the model. In \Cref{tab:ablations},  we train models with different input patch sizes but evaluate all of them on the same patch size of 32. Thus for the 64 and 128 patch size models, performance is evaluated on a $32\times32$ center crop of the model prediction.

Between the model trained with a patch size of 32 and the model trained with a patch size of 64, we see a consistent performance improvement in the $64\times64$ patch size model over the $32\times32$ patch size model.
This shows that performance degrades with a smaller context size and helps explain why Oya outperforms the other IR-based estimation products presented in this study, as they often use a much smaller context size(e.g., single pixels). 
However, increasing the input patch size to $128\times128$ ($640{km} \times 640{km}$) did not lead to similarly consistent gains over the $64\times64$ ($320{km} \times 320{km}$) patch size model. This suggests that, at the $5{km}$ resolution used, the $64\times64$ patches may already be sufficiently large to capture the dominant physical scales and contextual information of storm systems relevant for precipitation estimation.

\subsection{Impact of LDS}

To deal with the imbalance in the precipitation events in \Cref{section:handling-imbalance}, a loss re-weighting scheme based on the inverse of the label density estimated with LDS was used for training the QPE. We conduct an ablation and train a model without this loss-reweighting scheme to determine its effectiveness.
In \Cref{tab:ablations}, we compare the performance of two models, one trained without LDS and the other trained with LDS. We see that the LDS-based loss-reweighting scheme improves the accuracy of the model on heavy and extreme precipitation events as expected. Notably, it also remains comparable to the model trained without LDS on light and medium precipitation events.

% \begin{table}[h]
%     \small
%     \centering
%     \begin{tabular}{lcccc}
%     \toprule
%                   & \multicolumn{4}{c}{CSI}   \\ 
%         Retrieval &  Light & Medium & Heavy & Extreme  \\ 
%     \midrule
%         Oya (Without LDS) & 0.490 & \textbf{0.406} & 0.305 & 0.148 \\
%         Oya (With LDS) & \textbf{0.490} & 0.401 & \textbf{0.323} & \textbf{0.205} \\
%     \bottomrule
%     \end{tabular}
%     \caption{ \label{tab:lds} Effectiveness of LDS-based loss-reweighting scheme used in the regression model. Table shows the CSI of models trained with and without the loss-reweighting scheme. Applying the scheme leads to performance improvements on heavy and extreme precipitation.}
% \end{table}

\section{Global Estimation Dataset Generation}\label{section:dataset-generation}

Using the models described in this study, we designed an inference pipeline and generated with it a quasi-global precipitation estimation dataset from 2004 to present. The inference pipeline combines the outputs of the models trained for each GEO satellite or region to produce a consistent estimate covering latitudes $60 ^\circ$N-$60 ^\circ$S. In regions where the coverage of different GEO satellites overlaps, the estimates produced by the models for each satellite are averaged to produce a single, final estimate for each point. The historical record for the MSG coverage area was generated using the same models. This was possible due to the continuous availability of observations from 2004 to present. In contrast, the current generation of the other GEO satellites are not as extensive. Therefore, to construct a comparable long-term record for the Himawari region, we trained models using observations from the older Himawari-6 and Himawari-7 satellites, with precipitation observations from TRMM serving as the reference dataset. The dataset is made publicly available on Google Earth Engine at \url{https://developers.google.com/earth-engine/datasets/catalog/projects_global-precipitation-nowcast_assets_global_estimation}. 

\section{Conclusion}

This study introduces Oya, a novel deep learning algorithm for real-time, global precipitation estimation that leverages the full spectrum of visible and infrared channels from geostationary (GEO) satellites. To manage the significant data imbalance between rain and no-rain events, Oya uses a two-stage deep learning model. The model is trained using high-resolution GPM CORRA v07 data as ground truth and is pre-trained with IMERG-Final retrievals to enhance its robustness and prevent overfitting.

By using data from a constellation of GEO satellites, Oya achieves quasi-global coverage, producing estimates from $60 ^\circ$N-$60 ^\circ$S. We demonstrate that Oya significantly outperforms other GEO-based products like PDIR-Now and CRR across all tested precipitation intensities. Notably, Oya's performance is also superior to the near real-time IMERG Early product and is highly competitive with the research-grade IMERG Final product, which is produced with a 3.5-month latency. Ablation studies confirmed the benefits of key design choices, including pre-training, the use of all available GEO channels, and the application of Label Distribution Smoothing (LDS) to handle data imbalances.

The creation of a publicly available quasi-global precipitation dataset from this work provides a valuable new resource for the scientific community. While challenges remain, such as mitigating parallax and limb darkening effects at higher latitudes, the Oya algorithm represents a significant advancement in satellite-based precipitation estimation, with the potential to greatly benefit applications in agriculture, water resource management, and disaster preparedness.

We would like to thank Shreya Agrawal for contributions to this research.

\bibliographystyle{tmlr}
\bibliography{refs}
\vfill

\renewcommand{\thesection}{\Alph{section}}
\setcounter{section}{0}

\pagebreak

\section{Appendix}

\subsection{Spectral channels (in microns) available on the current generation of GEO satellites }\label{section:channels-summary}
\begin{table}[h]

\begin{tabular}{llccc}
 \toprule
Category      & Channel Description                     & Meteosat & \multicolumn{1}{l}{GOES} & \multicolumn{1}{l}{Himiwari} \\ \midrule
Visible Bands & Blue                                 &          & 0.47                      & 0.47                          \\ 
              & Green                                &          &                           & 0.51                          \\ 
              & Red                                  & 0.6      & 0.64                      & 0.64                          \\ \bottomrule
Near IR Bands & Vegetation                               & 0.8      & 0.86                      & 0.86                          \\ 
              & Cirrus                               &          & 1.37                      & \multicolumn{1}{l}{}         \\ 
              & Snow/Ice                             & 1.6      & 1.6                       & 1.6                           \\ 
              & Cloud Particle Size                  &          & 2.2                       & 2.3                           \\ \bottomrule
IR Bands      & Shortwave Window                     & 3.9      & 3.9                       & 3.9                           \\ 
              & Upper-Level Tropospheric Water Vapor & 6.2      & 6.2                       & 6.2                           \\ 
              & Mid-Level Tropospheric Water Vapor   &          & 6.9                       & 6.9                           \\ 
              & Lower-level Water Vapor              & 7.3      & 7.3                       & 7.3                           \\ 
              & Cloud-Top Phase                      & 8.7      & 8.4                       & 8.6                           \\ 
              & Ozone Band                           & 9.7      & 9.6                       & 9.6                           \\
            %   \bottomrule
              & Clean                                &          & 10.3                      & 10.4                          \\ 
              & Longwave Window                      & 10.8     & 11.2                      & 11.2                          \\ 
              & Dirty                                & 12.0     & 12.3                      & 12.4                          \\ 
              & CO2                                  & 13.4     & 13.3                      & 13.3                          \\ \bottomrule
\end{tabular}
\caption{ \label{tab:band-summary} Summary comparison of the channels available on the current generation of GEO satellites.}
\end{table}

\subsection{Definition of Metrics}\label{section:metric-definitions}

Given a precipitation threshold $T$, we compute the number of pixels correctly classified or misclassified as being greater than or equal to $T$.
This gives us the true positives (TP), true negatives (TN), false positives (FP) and false negatives (FN) from which the metrics are computed as follows:

\begin{equation}
    CSI = \frac{TP}{TP + FP + FN}
\label{eq:csi}
\end{equation}

\begin{equation}
    POD = \frac{TP}{TP + FN}
\label{eq:pod}
\end{equation}

\begin{equation}
    Bias = \frac{TP + FP}{TP + FN}
\label{eq:bias}
\end{equation}

\begin{equation}
    FAR = \frac{FP}{TP + FP}
\label{eq:far}
\end{equation}

The following thresholds are used for the evaluations presented in this paper: 0.2 $mm\:h^{-1}$ (light precipitation), 1.0 $mm\:h^{-1}$ (medium precipitation), 2.4 $mm\:h^{-1}$ (heavy precipitation), 7.0 $mm\:h^{-1}$ (extreme precipitation) (closely matching levels in the AMS Glossary of Meteorology~\citep{ams-glossary-rain}).

% Given a precipitation threshold $T$, these metrics compute the ratio of pixels correctly classified or misclassified as being greater than or equal to $T$ to those that were not \citep{CreatingSyntheticRadarImageryUsingConvolutionalNeuralNetworks}.

\end{document}